\documentclass[runningheads]{llncs}
\usepackage{hyperref}

\usepackage{graphicx}
\usepackage{caption}
\usepackage[position=top]{subfig}
\usepackage{amsmath,amssymb} 

\usepackage[width=122mm,height=193mm,centering]{geometry}

\usepackage{color}
\usepackage{xspace}
\usepackage{booktabs}
\usepackage{tabulary}
\usepackage{floatrow}
\usepackage{wrapfig}
\usepackage[demo,abs]{overpic}

\newfloatcommand{capbtabbox}{table}[][\FBwidth]

\makeatletter
\newcommand*{\etc}{%
  \@ifnextchar{.}%
    {etc}%
    {etc.\@\xspace}%
}
\makeatother
\newcommand{\etal}{et al.\@\xspace}
\newcommand{\eg}{e.g.\@\xspace}
\newcommand{\ie}{i.e.\@\xspace}

\newcommand{\vs}{vs.\@\xspace}
\newcommand{\synsets}{\mathcal{S}}
\newcommand{\ig}{\mathcal{I}}
\newcommand{\ighashtags}{\mathcal{H}}
\newcommand{\scinum}[2]{{#1}\mathrm{e}{#2}}
\newcommand{\app}{\raise.17ex\hbox{$\scriptstyle\sim$}}
\makeatletter\renewcommand\paragraph{\@startsection{paragraph}{4}{\z@}
  {.5em \@plus1ex \@minus.2ex}{-.5em}{\normalfont\normalsize\bfseries}}\makeatother

\newif\ifcomments

    \commentstrue

\ifcomments
\newcommand{\comments}[1]{#1}

\else
\newcommand{\comments}[1]{\ignorespaces}

\fi

\newcommand*{\plottitle}{\bf\fontfamily{phv}\fontsize{7}{9}\selectfont}

\newlength\savewidth
\newcommand{\tablestyle}[2]{\setlength{\tabcolsep}{#1}\renewcommand{\arraystretch}{#2}\centering\footnotesize}
\newcolumntype{x}[1]{>{\centering\arraybackslash}p{#1pt}}

\begin{document}
\pagestyle{headings}
\mainmatter
\def\ECCV18SubNumber{1659}  

\title{Exploring the Limits of\\Weakly Supervised Pretraining}
\titlerunning{Exploring the Limits of Weakly Supervised Pretraining}


\authorrunning{Mahajan \etal}
\author{Dhruv Mahajan \qquad Ross Girshick \qquad Vignesh Ramanathan \qquad Kaiming He \\ Manohar Paluri ~~~ Yixuan Li ~~~ Ashwin Bharambe ~~~ Laurens van der Maaten}
\institute{Facebook\vspace{-1em}}
\maketitle

\begin{abstract}
State-of-the-art visual perception models for a wide range of tasks rely on supervised pretraining. ImageNet classification is the de facto pretraining task for these models. Yet, ImageNet is now nearly ten years old and is by modern standards ``small''. Even so, relatively little is known about the behavior of pretraining with datasets that are multiple orders of magnitude larger. The reasons are obvious: such datasets are difficult to collect and annotate. In this paper, we present a unique study of transfer learning with large convolutional networks trained to predict hashtags on \emph{billions} of social media images. Our experiments demonstrate that training for large-scale hashtag prediction leads to excellent results. We show improvements on several image classification and object detection tasks, and report the highest ImageNet-1k single-crop, top-1 accuracy to date: 85.4\% (97.6\% top-5). We also perform extensive experiments that provide novel empirical data on the relationship between large-scale pretraining and transfer learning performance.
\end{abstract}

\section{Introduction}
\label{sec:introduction}

Nearly all state-of-the-art visual perception algorithms rely on the same formula: (1) pretrain a convolutional network on a large, manually annotated image classification dataset and (2) finetune the network on a smaller, task-specific dataset. This formula \cite{Girshick2014,Donahue2013,Zeiler2014} has been in wide use for several years and led to impressive improvements on numerous tasks. Examples include: object detection \cite{Girshick2014,he2017mask}, semantic segmentation \cite{Long2015,Zhao2017}, human pose estimation \cite{Cao_2017_CVPR,Papandreou_2017_CVPR}, video recognition \cite{Carreira_2017_CVPR}, monocular depth estimation \cite{Eigen2015}, and so on. In fact, it is so effective that it would now be considered foolhardy \emph{not} to use supervised pretraining.

The ImageNet dataset \cite{russakovsky2015} is the de facto pretraining dataset. While there are studies analyzing the effects of various ImageNet pretraining factors on transfer learning (\eg, \cite{Agrawal2014,huh2016}) or the use of different datasets that are of the same size magnitude as ImageNet (\eg, \cite{zhou2017places,xie2017resnext}), relatively little is known about pretraining on datasets that are \emph{multiple} orders of magnitude larger (\cite{joulin2016learning,sun2017unreasonable} are the largest studies to date). The reasons for this are numerous: few such datasets exist, building new datasets is labor intensive, and large computational resources are needed to conduct experiments. Yet, given the central role of pretraining it is important to expand our scientific knowledge in this domain.

This paper tries to address this complex issue by studying an unexplored data regime: \emph{billions of images ``labeled'' in the wild with social media hashtags}. This data source has the advantage of being large and continuously growing, as well as ``free'' from an annotation perspective since no manual labeling is required. However, the data source also has potential disadvantages: hashtags may be too noisy to serve as an effective supervisory signal and the image distribution might be biased in ways that harm transfer learning. It is not a priori obvious that training on this data will yield good transfer learning results.

The main result of this paper is that without manual dataset curation or sophisticated data cleaning, models trained on billions of Instagram images using thousands of distinct hashtags as labels exhibit excellent transfer learning performance. For example, we observe improvements over the state-of-the-art for image classification and object detection, where we obtain a single-crop, top-1 accuracy of 85.4\% on the ImageNet-1k image-classification dataset and 45.2\% AP on the COCO object-detection dataset \cite{mscoco2014}, compared to 79.8\% and 43.7\%, respectively, when training (or pretraining) the same models on ImageNet-1k. Our primary goal, however, is to contribute novel experimental data about this previously unexplored regime. To that end, we conduct numerous experiments that reveal interesting trends. For example, we find that ``hashtag engineering'' (\ie, collecting images tagged with a specific subset of hashtags) is a promising new direction for improving transfer learning results, that training on large-scale hashtag data is unexpectedly robust to label noise, and that the features learned allow a simple linear classifier to achieve state-of-the-art ImageNet-1k top-1 accuracy of 83.6\% without any finetuning (compared to 84.2\% with finetuning).

\section{Scaling up Supervised Pretraining}
\label{sec:methods}

In our experiments, we train standard convolutional network architectures to predict hashtags on up to 3.5 billion public Instagram images. To make training at this scale practical, we adopt a distributed synchronous implementation of stochastic gradient descent with large (8k image) minibatches, following Goyal \etal \cite{goyal2017accurate}. We experiment on a variety of datasets, which we describe next.

\subsection{Instagram Datasets}
\label{sec:ig_datasets}

We use a simple data collection pipeline: (1) We select a set of hashtags. (2) We download images that are tagged with at least one of these hashtags. (3) Then, because multiple hashtags may refer to the same underlying concept, we apply a simple process that utilizes WordNet \cite{wordnet} synsets to merge some hashtags into a single canonical form (\eg, \texttt{\#brownbear} and \texttt{\#ursusarctos} are merged). (4) Finally, for each downloaded image, we replace each hashtag with its canonical form and discard any hashtags that were not in the selected set. The canonical hashtags are used as labels for training and evaluation.

By varying the selected hashtags and the number of images to sample, we can construct a variety of datasets of different sizes and visual distributions. Table \ref{tab:datasets} summarizes the datasets used in our experiments. Each dataset is named by completing a template, \emph{role}-\emph{source}-$I$-$L$, that indicates its role (training, validation, testing), source (IG for Instagram, IN for ImageNet, \etc), number of images $I$, and number of labels $L$. We use approximate image and label counts for convenience, for example ``train-IG-940M-1.5k'' is an Instagram dataset for training with \app $\scinum{940}{6}$ images and \app 1,500 labels. We omit the role and image count when it is clear from context or not useful to present.

We design three hashtag sets for the Instagram data: (1) A \app 1.5k set with hashtags from the standard 1,000 IN-1k synsets (each synset contains at least one synonym, hence there are more hashtags than synsets). (2) A \app 17k set with hashtags that are synonyms in any of the noun synsets in WordNet. And (3) an \app 8.5k set with the most frequent hashtags from the 17k set. The hashtag set sizes are measured after merging the hashtags into their canonical forms. We hypothesize that the first set has a visual distribution similar to IN-1k, while the other two represent more general visual distributions covering fine-grained visual categories. Details of how these hashtags are selected and how the merging process works are given in supplemental material.

\begin{table*}[t]
\caption{\small \textbf{Summary of image classification datasets.} Each dataset is named with a template, \emph{role}-\emph{source}-$I$-$L$, that indicates its role (training, validation, testing), source, number of images $I$, and number of labels $L$.}
\resizebox{0.91\linewidth}{!}{
{\tablestyle{4pt}{1.1}
\begin{tabular}{lll}
\toprule
\bf Name template & \bf Description \\
\midrule
  train-IG-$I$-1.5k  & Instagram training set of $I$ images and \app 1.5k hashtags from ImageNet-1k. \\
  train-IG-$I$-8.5k  & Instagram training set of $I$ images and \app 8.5k hashtags from WordNet. \\
  train-IG-$I$-17k   & Instagram training set of $I$ images and \app 17k hashtags from WordNet. \\
\hline
  train-IN-1M-1k  & The standard ImageNet-1k ILSVRC training set with 1.28M images. \\
  val-IN-50k-1k     & The standard ImageNet-1k ILSVRC validation set with  50k images.\\
\hline
  train-IN-$I$-$L$  & Extended ImageNet training set of $I$ images and $L \in \{5\mathrm{k}, 9\mathrm{k}\}$ labels. \\
  val-IN-$I$-$L$    & Extended ImageNet validation set of $I$ images and $L \in \{5\mathrm{k}, 9\mathrm{k}\}$ labels. \\
\hline
  train-CUB-6k-200 & The Caltech-UCSD Birds-200-2011 training set. \\
  val-CUB-6k-200   & The Caltech-UCSD Birds-200-2011 validation set. \\
\hline
  train-Places-1.8M-365 & The Places365-Standard training set (high-resolution version). \\
  val-Places-37k-365    & The Places365-Standard validation set (high-resolution version). \\
\hline
  train-COCO-135k-80 & The standard COCO detection training set (2017 version). \\
  val-COCO-5k-80 & The standard COCO detection validation set (2017 version). \\
  test-COCO-20k-80 & The standard COCO detection test-dev set (2017 version). \\
\bottomrule
\end{tabular}}
}
\vspace{-0.5em}
\label{tab:datasets}
\end{table*}


\paragraph{Image deduplication.} When performing transfer learning, it is essential to understand and properly address overlap between training and test sets. Overlap can exists because images may come from the same underlying sources (\eg, Wikipedia, Flickr, Google). For instance, $\app 5\%$ of the images in the val-CUB-6k-200 set \cite{welinder2010birds} also appear in train-IN-1M-1k, and $1.78\%$ of images in val-IN-50k-1k set are in the JFT-300M training set \cite{sun2017unreasonable}. To address this issue, we performed the following deduplication procedure: we compute R-MAC features \cite{gordo2016rmac,tolias2016rmac} for all candidate images using a ResNet-50 model, and use these features to find the $k\!=\!21$ nearest neighbors for each of the images in our test sets (additional details are in the supplemental material). Subsequently, we manually inspected all images and their nearest neighbors to identify duplicates. This procedure uncovered $150$ val-IN-50k-1k (0.30\%), $10$ val-CUB-6k-200 (0.17\%), $151$ val-Places-37k-365 ($0.41\%$), and $6$ val-COCO-5k-80 ($0.12\%$) duplicates. In our results, we report the observed accuracy of our models; in the supplemental material, we report a conservative lower bound on accuracy by marking all duplicates as incorrect. Given the small percentage of duplicates, they do not impact our findings.

\paragraph{Discussion.} Our datasets have two nice properties: public visibility and simplicity. By using publicly accessible images, the data used in our experiments is visible to everyone. To see what it looks like, the images are browsable by hashtag at \texttt{https://www.instagram.com/explore/tags/} followed by a specific hashtag; for example \url{https://www.instagram.com/explore/tags/brownbear} shows images tagged with \texttt{\#brownbear}. Our data is also taken from the ``wild'', essentially as-is, with minimal effort to sanitize it. This makes the dataset construction process particularly simple and transparent.

We contrast these properties with the JFT-300M dataset \cite{sun2017unreasonable}, which is not publicly visible and is the result of a proprietary collection process (``The [JFT-300M] images are labeled using an algorithm that uses a complex mixture of raw web signals, connections between web-pages and user feedback.''). Additional details describing the collection of JFT-300M have not been publicly disclosed.

Despite our efforts to make the dataset content and collection process transparent, we acknowledge that, similar to JFT-300M, it is not possible for other research groups to know exactly which images we used nor to download them en masse. Hence it is not possible for others to replicate our results at this time. However, we believe that it is better if we undertake this study and share the results with the community than to not publish the results.

\subsection{ImageNet Datasets}

In addition to the standard IN-1k dataset, we experiment with larger subsets of the full ImageNet 2011 release that contains 14.2M images and 22k labels. We construct training and validation sets that include 5k and 9k labels. For the 5k set, we use the now standard IN-5k proposed in \cite{xie2017resnext} (6.6M training images). For the 9k label set, we follow the same protocol used to construct IN-5k, which involves taking the next most frequent 4k labels and all of the associated images (10.5M training images). In all cases, we use 50 images per class for validation.

\subsection{Models}
We use residual networks with grouped convolutional layers, called ResNeXt \cite{xie2017resnext}. Our experiments use ResNeXt-101 32$\times C$d, which has 101 layers, 32 groups, and group widths $C$ of: 4 (8B multiply-add FLOPs, 43M parameters), 8 (16B, 88M), 16 (36B, 193M), 32 (87B, 466M), and 48 (153B, 829M). Our implementation matches \cite{goyal2017accurate}. We believe our results will generalize to other architectures \cite{he2016deep,huang2017densenet,szegedy2016inceptionv4}.

\paragraph{Loss function.} In contrast to ImageNet, our Instagram datasets may contain multiple labels per image (because a user specified multiple hashtags). The average number of hashtags per image varies depending on the dataset; for instance, train-IG-1B-17k contains \app 2 hashtags per image. Our model computes probabilities over all hashtags in the vocabulary using a softmax activation and is trained to minimize the cross-entropy between the predicted softmax distribution and the target distribution of each image. The target is a vector with $k$ non-zero entries each set to $1/k$ corresponding to the $k \ge 1$ hashtags for the image.

We have also experimented with per-hashtag sigmoid outputs and binary logistic loss, but obtained significantly worse results. While counter-intuitive given the multi-label data, these findings match similar observations in \cite{joulin2016learning}. The successful application of sigmoid activations and logistic loss may require sophisticated label completion techniques \cite{sun2017unreasonable} and more hyper-parameter search.

\subsection{Pretraining Details}

Our models are trained by synchronous stochastic gradient descent (SGD) on 336 GPUs across 42 machines with minibatches of 8,064 images. Each GPU processes 24 images at a time and batch normalization (BN) \cite{Ioffe2015} statistics are computed on these 24 image sets. The length of the training schedule, measured in units of number-of-images-processed (\ie, minibatch size $\times$ total SGD updates), is determined by a heuristic: we choose two training extremes (for instance, $120$ epochs on $\scinum{1.2}{6}$ images and $2$ epochs on $\scinum{3.5}{9}$ images) and linearly interpolate the schedule between them to set the number-of-images-processed for each experiment. Schedules for each experiment are in the supplemental material. Our ResNeXt-101 32$\times$16d networks took $\app 22$ days to train on 3.5B images.

To set the learning rate, we follow the linear scaling rule with gradual warm-up described in \cite{goyal2017accurate}. We use a warm-up from 0.1 up to $0.1 / 256 \times 8064$, where 0.1 and 256 are canonical learning rate and minibatch sizes \cite{He2015}. After the warm-up, the learning rate is multiplied by 0.5 at equally spaced steps, such that the total number of learning rate reductions is 20 over the course of training. The same settings are used when training on ImageNet and Instagram data, except that when training on ImageNet we use 128 GPUs in 16 machines (for a minibatch size of 3,072) due to the smaller dataset size and we use the standard learning rate schedule that involves three equally spaced reductions by a factor of 0.1. All other initialization and training details match \cite{goyal2017accurate} and are summarized in the supplemental material.

\section{Experiments}
\label{sec:experiments}

In our experiments, we pretrain convolutional networks for hashtag prediction and transfer those networks to a variety of tasks. There are two established protocols for judging the quality of a pretrained model (see \cite{pathak2016motion} \S3 for a discussion). Both analyze how pretraining on a \emph{source task}, \eg IN-1k classification, leads to gains (or losses) on a \emph{target task}, \eg bird recognition or object detection.

\emph{Full network finetuning} views pretraining as sophisticated weight initialization: the success of pretraining is judged by its impact on the target task after further training the network weights in a task-specific manner (\ie finetuning). By contrast, \emph{feature transfer} uses the pretrained network as a feature extractor: it judges the quality of the network by how effective its features are on other tasks, without updating any of the network parameters. These protocols are two extremes of a spectrum along which the proportion of pretrained weights that are finetuned varies from all to none. We employ both protocols in our experiments; at times one is more appropriate than the other.

\paragraph{Full network finetuning} is performed by removing the hashtag-specific fully connected classification layer from the network and replacing it with a randomly initialized classification layer with one output per class in the target task. This modified network is then trained using SGD with momentum. We select the finetuning learning rate and schedule by grid search on a proper validation set for each target task. To do this, we randomly hold out a small portion of the \emph{training} set (see supplemental material). This practice ensures that our results on the standard validation sets are clean.

\paragraph{Feature transfer} is performed by training an L2-regularized linear logistic regressor on the training data for the target task using SGD. The features produced by the pretrained network are used as input into the classifier. We train the classifier until convergence to the global optimum.

\subsection{Image Classification Experiments}
We evaluate Instagram pretraining by measuring classification accuracies on three classification target tasks: ImageNet \cite{deng2009imagenet}, CUB2011 \cite{welinder2010birds}, and Places365 \cite{zhou2017places}. We perform inference on 224$\times$224 center-cropped images, and study the effects of (1) the hashtag vocabulary size, (2) the training set size, (3) the amount of noise in the hashtag targets, and (4) the hashtag sampling strategy.

\paragraph{3.1.1~~How does the Instagram hashtag set impact accuracy?} Our first experiment varies the Instagram hashtag sets used in pretraining (1.5k, 8.5k, \vs 17k) whilst keeping other factors constant. We compute transfer learning results as top-1 classification accuracy on five target datasets: val-IN-1k, val-IN-5k, val-IN-9k, val-CUB-200, val-Places-365. For baseline models, we use ImageNet classification as a source task: we train networks on train-IN-1k, train-IN-5k, and train-IN-9k, and evaluate them on the corresponding validation sets (finetuning is not needed in these cases). For val-CUB-200 and val-Places-365, we use train-IN-1k as the baseline source task and finetune on train-CUB-200 and train-Places-365. Full network finetuning of ResNeXt-101 32$\times$16d is used for all source-target pairs in which source and target are not the same.

\begin{figure}[t!]
\centering
\subfloat[{\plottitle ~~~Target task: ImageNet}]{\includegraphics[width=0.5\linewidth]{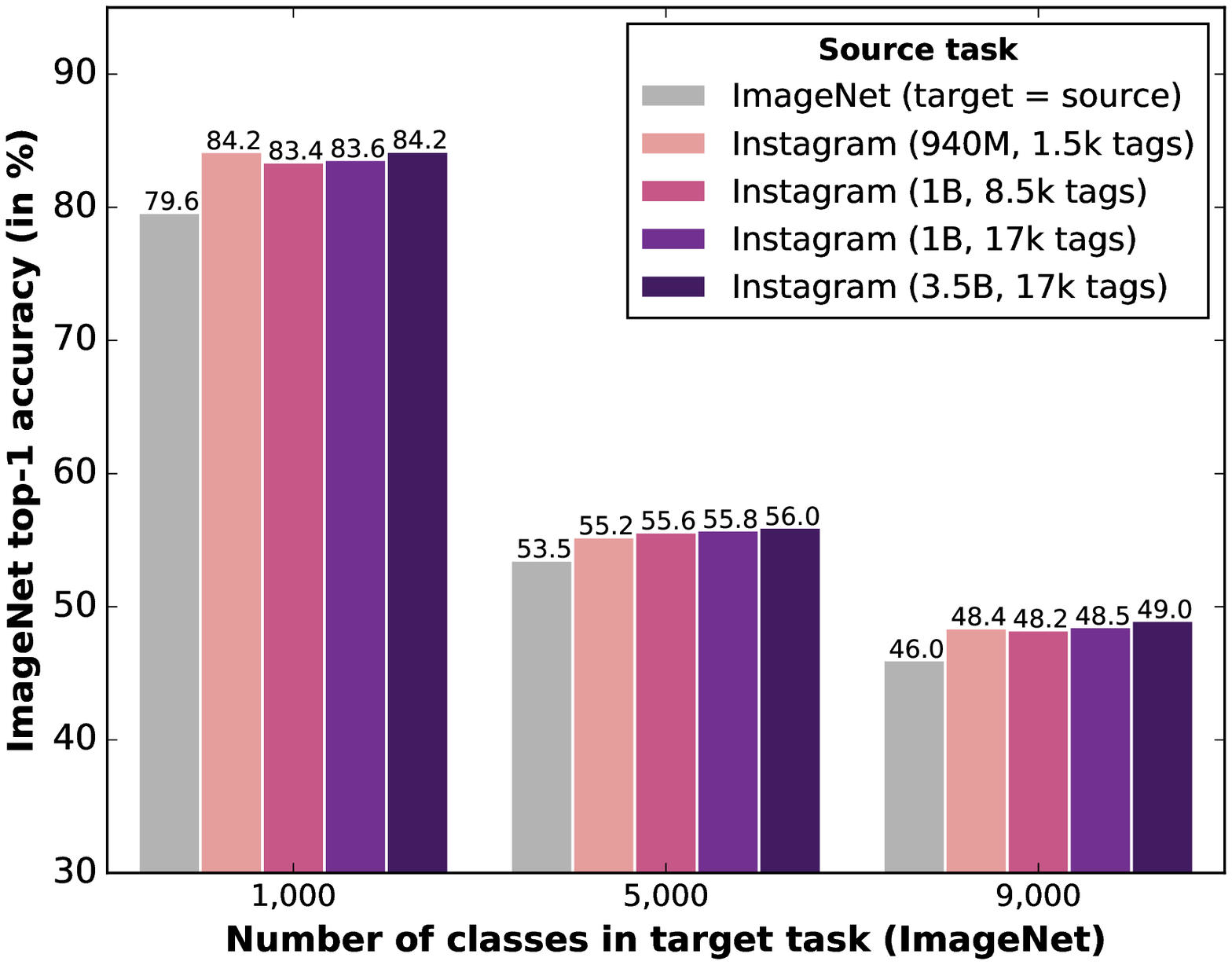}}%
\subfloat[{\plottitle ~~~Target task: CUB \& Places}]{\includegraphics[width=0.5\linewidth]{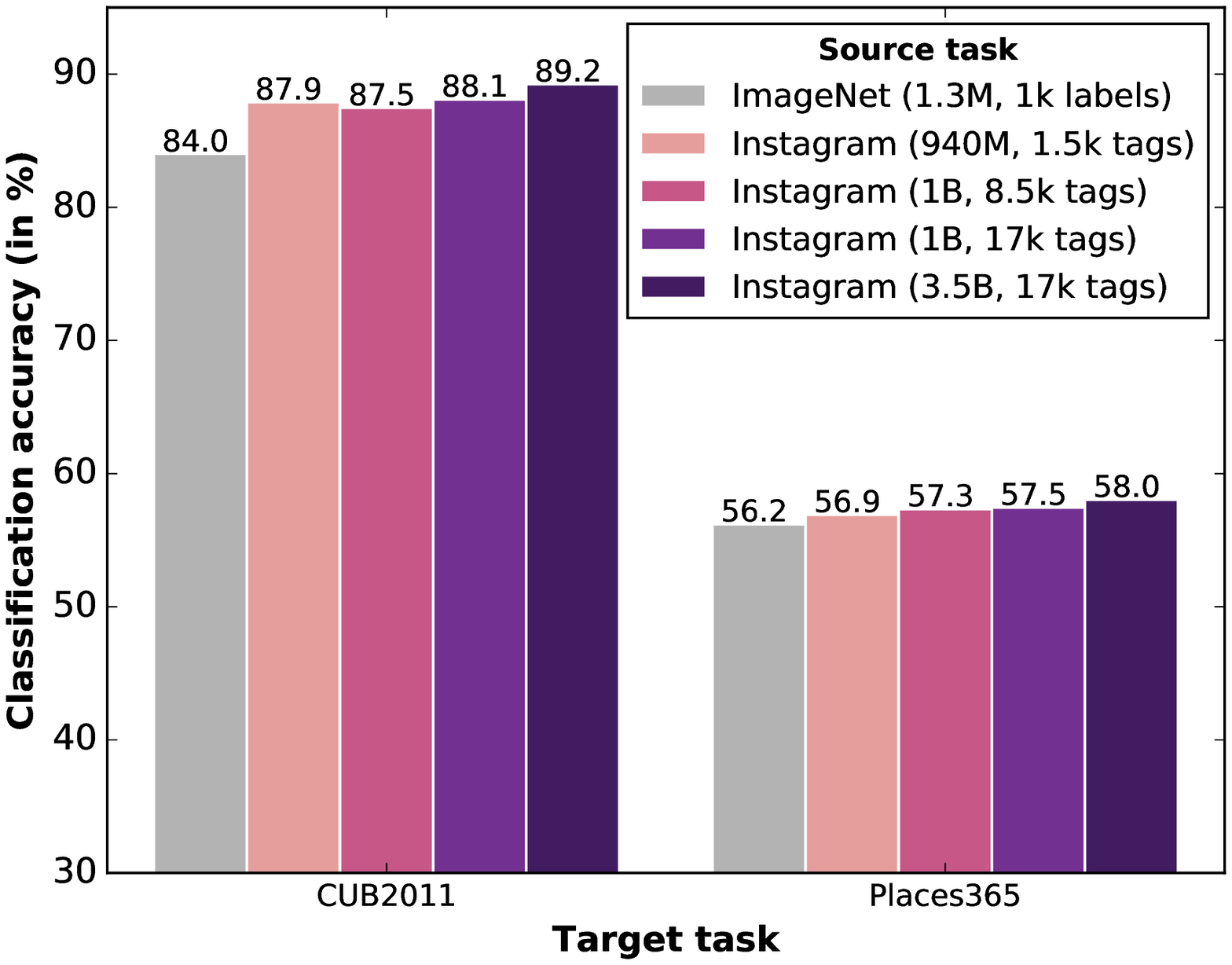}}%
\vspace{-0.5em}%
\caption{\small Classification accuracy of ResNeXt-101 32$\times$16d pretrained on IG-1B with different hashtag vocabularies (purple bars) on IN-\{1k, 5k, 9k\} (left) and CUB2011, Places365 (right). Baseline models (gray bars) are trained on IN-\{1k, 5k, 9k\} (left) and IN-1k (right), respectively. Full network finetuning is used. Higher is better.}
\label{fig:figure1}
\end{figure}

Figure~\ref{fig:figure1} shows that pretraining for hashtag prediction substantially improves target task accuracy: on the standard IN-1k benchmark set, a network pretrained on nearly 1B Instagram images with 1.5k hashtags achieves a state-of-the-art accuracy of 84.2\%---an improvement of 4.6\% over the same model architecture trained on IN-1k alone and a 1.5\% boost over the prior state-of-the-art \cite{zoph2017transferable}, which uses an optimized network architecture. The performance improvements due to Instagram pretraining vary between ImageNet tasks: on the 1k class task, the model pretrained with the IN-1k-aligned 1.5k hashtag set outperforms source networks trained on larger hashtag sets. This trend reverses as the number of target ImageNet classes increases: on 9k ImageNet target classes, the model pretrained with 17k hashtags strongly outperforms the 1.5k hashtag model. On the CUB2011 and Places365 target tasks, source models trained with the largest hashtag sets perform the best, likely, because the 17k hashtags span more objects, scenes, and fine-grained categories. These patterns are intuitive and suggest that alignment between the source and target label sets is an important factor.

We also show results in Figure~\ref{fig:figure1} using a larger 3.5B image set with 17k hashtags (dark purple bars), which performs best across all target tasks. Furthermore, following \cite{storck2017convnets}, we measure the \emph{rectified} classification accuracy of this model on val-IN-1k. We present all incorrect classifications to five human annotators, asking whether or not the prediction is correct: if at least four annotators answer this question affirmatively the model's prediction is considered correct. Whereas the IN-1M-1k model obtained a rectified top-1 accuracy of $87.5\%$ on val-IN-1k, our IG-3.5B-17k pretrained model achieved $90.4\%$.

\begin{figure}[h!]
  \centering
  \subfloat[{\plottitle ~~~~Target task: ImageNet-1k}]{\includegraphics[width=0.5\linewidth]{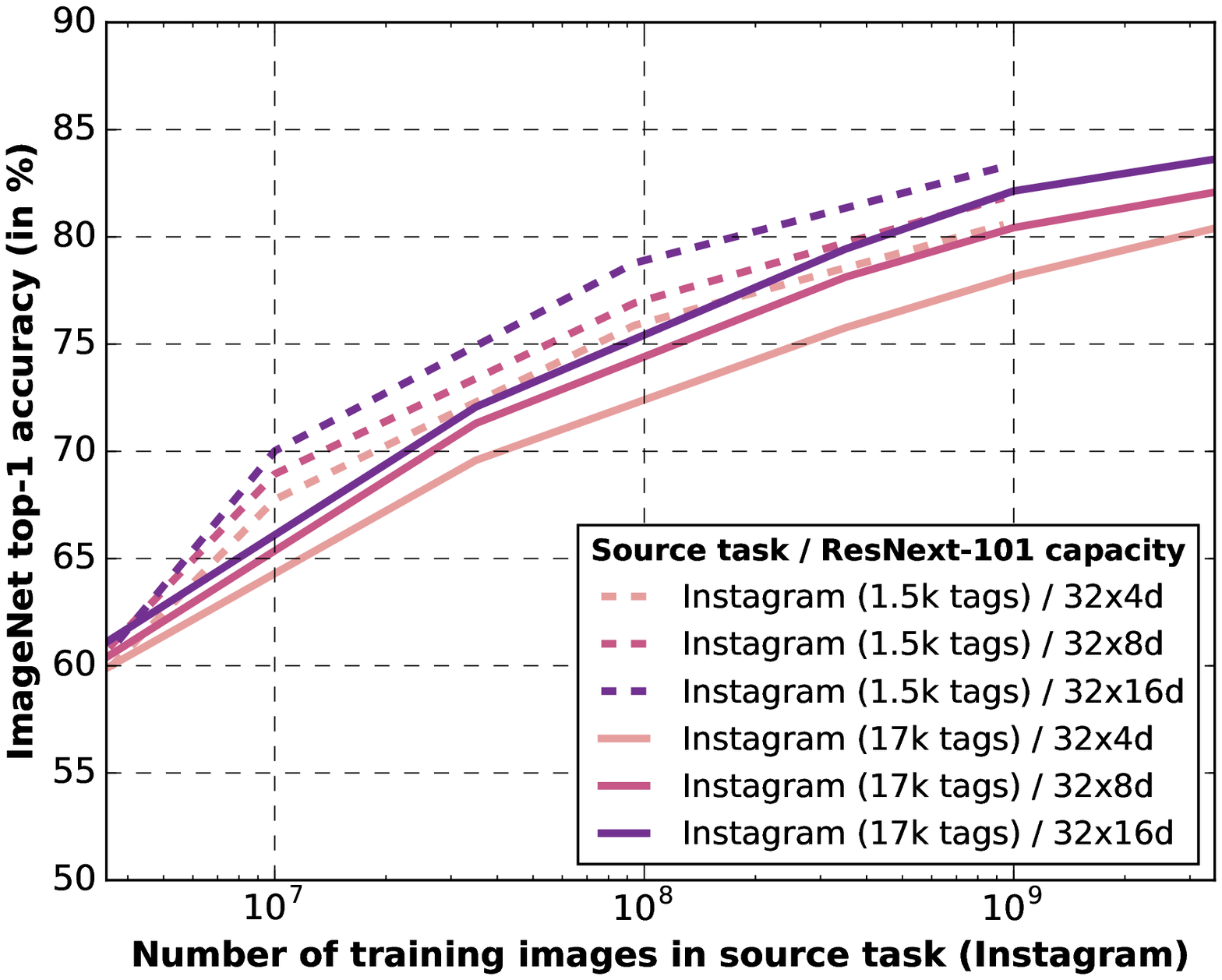}}
  \subfloat[{\plottitle ~~~~Target task: ImageNet-5k}]{\includegraphics[width=0.5\linewidth]{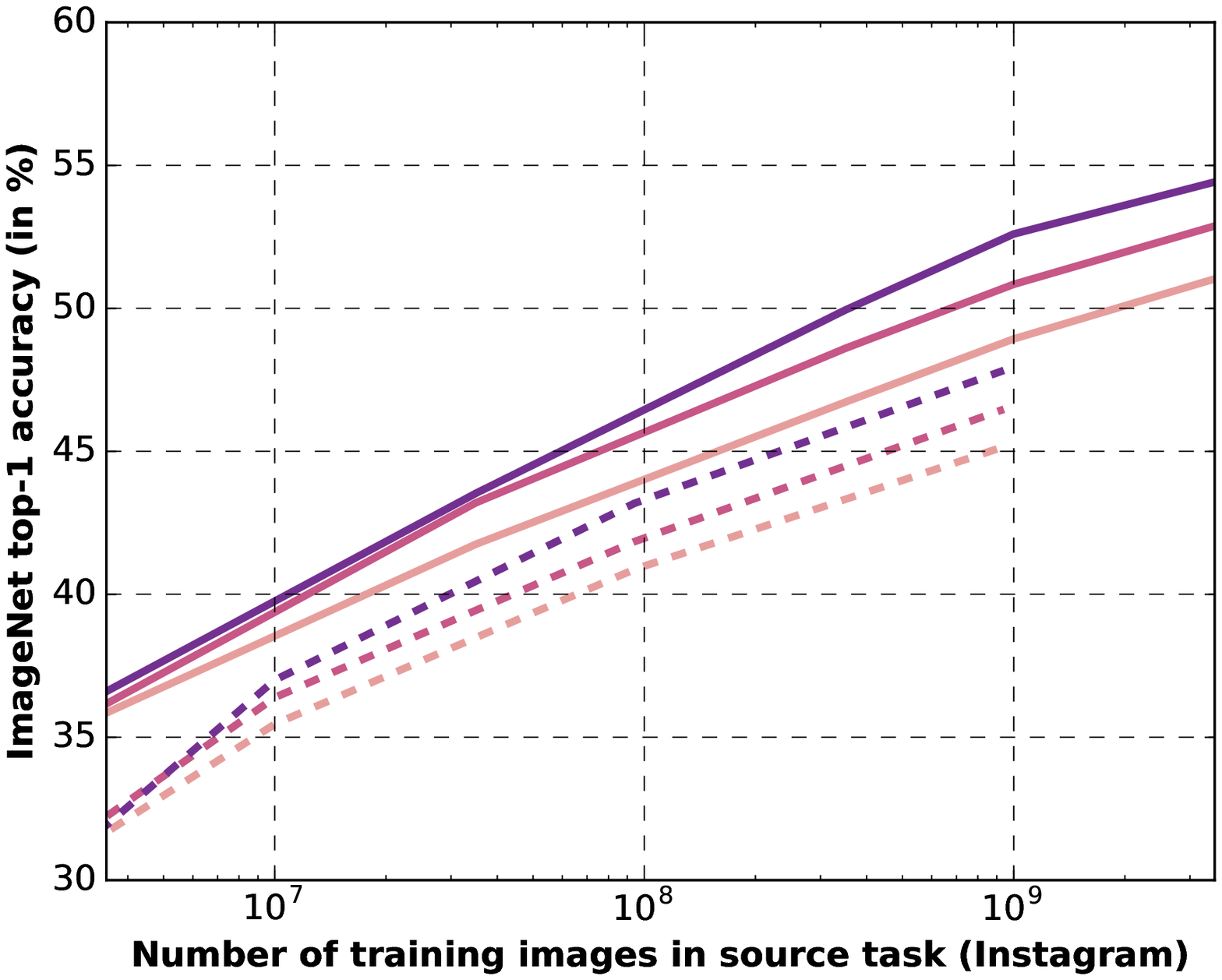}}\\
  \vspace{-0.5em}
  \subfloat[{\plottitle ~~~~Target task: ImageNet-9k}]{\includegraphics[width=0.5\linewidth]{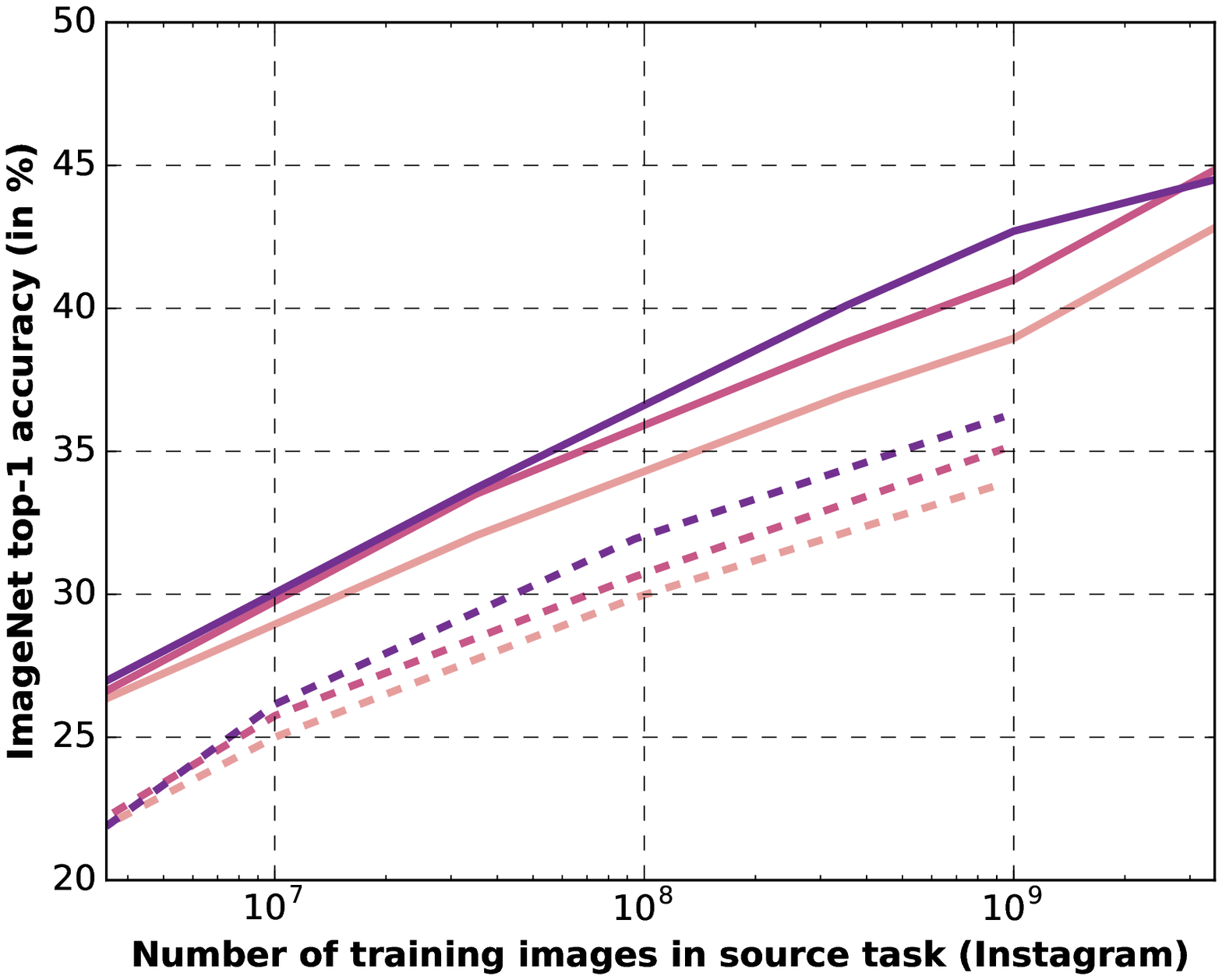}}
  \subfloat[{\plottitle ~~~~Target task: CUB2011}]{\includegraphics[width=0.5\linewidth]{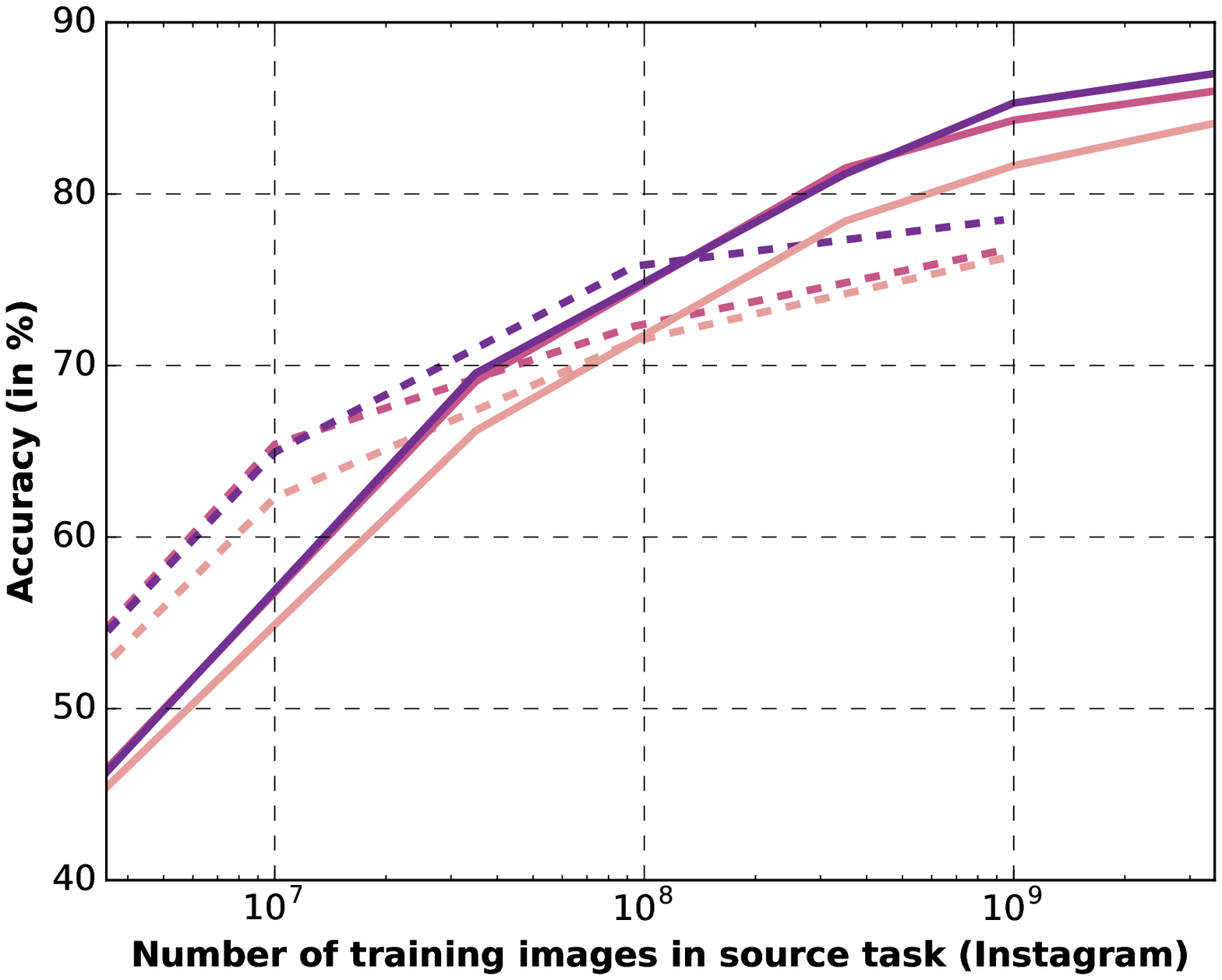}}
  \vspace{-0.5em}
  \caption{\small Classification accuracies on IN-\{1k, 5k, 9k\} and CUB2011 target tasks as a function of the number of Instagram images used for pretraining for three network architectures (colors) and two hashtag vocabularies (dashed / solid lines). Only the linear classifier is trained on the target task. Higher is better.}
\label{fig:figure2}
\end{figure}

\paragraph{3.1.2~~How does the pretraining image set size impact accuracy?} This experiment studies the relationship between the number of images used in Instagram pretraining and classification accuracy on the target task. For these experiments, when transferring to the target task we keep the pretrained network weights fixed and only train a linear classifier for the target task. We make this choice because when the number of pretraining images is small relative to the number of target task images (\eg, 1M \vs 7M), the effect of pretraining is masked by the large amount of finetuning data (this was not the case in the previous experiment where the source task had orders of magnitude more images).

Figure~\ref{fig:figure2} shows the classification accuracy on ImageNet validation sets ($y$-axis) as a function of the number of Instagram training images ($x$-axis; note the log scale) ranging from 3.5M to 3.5B images. The figure shows results for models pretrained to predict 1.5k hashtags (dashed lines) or 17k hashtags (solid lines) for ResNeXt-101 models with three different capacities (represented by different colors).\footnote{The maximum number of images available for the 1.5k hashtag set is 940M.} The four panels correspond to ImageNet target tasks with three different number of classes (1k, 5k, 9k) and CUB2011.

In line with prior results \cite{joulin2016learning,sun2017unreasonable}, we observe near log-linear behavior: each time we multiply the amount of training data by a factor of $x$, we observe a fixed increase $y$ in classification accuracy. While the scaling behavior is consistent across hashtag vocabulary sizes and models, the accuracy increase $y$ is larger for higher-capacity networks: across all figures, the lines corresponding to ResNeXt-101 32$\times$16d networks (purple) are steeper than those corresponding to 32$\times$8d and 32$\times$4d models. This result suggests that when training convolutional networks on billions of training images, current network architectures are prone to underfitting. We also observe log-linear scaling break down in two regimes: (1) because accuracy is bounded, endless log-linear scaling is not possible. On datasets like IN-1k and CUB2011 the ceiling effect necessarily creates sub-log-linear scaling. (2) We observe a deviation from log-linear scaling in the 1B to 3.5B image regime even without apparent ceiling effects on IN-\{5k, 9k\}.

These plots also illustrate an interesting effect of the hashtag vocabulary on the transfer task accuracy. On IN-1k, networks pretrained on the target-task-aligned 1.5k hashtags outperform those trained using a larger hashtag vocabulary, because the 1.5k hashtags were selected to match the ImageNet synsets. However, as the matching between hashtag vocabulary and target classes disappears and the visual variety in the transfer task increases, networks pretrained to recognize a larger number of hashtags increasingly outperform networks pretrained on fewer hashtags: on the IN-9k transfer task, the difference in accuracy between networks trained on 1.5k and those trained on 17k hashtags is $\app 7\%$.

The highest accuracies on val-IN-1k are 83.3\% (source: IG-940M-1k) and 83.6\% (source: IG-3.5B-17k), both with ResNeXt-101 32$\times$16d. \emph{These results are obtained by training a linear classifier on fixed features and yet are nearly as good as full network finetuning, demonstrating the effectiveness of the feature representation learned from hashtag prediction.} These results also have low variance: we pretrained the ResNeXt-101 32$\times$16d architecture of two different random samples of 1B images and then trained linear classifiers on IN-\{1k, 5k, 9k\} finding a difference in top-1 accuracy of less than 0.1\% in all cases.

To test whether the above observations generalize to fine-grained classification, we repeated the experiments on the CUB2011 dataset, and show the results in Figure~\ref{fig:figure2}, bottom right. The curves reveal that when training data is limited, the 1.5k hashtag dataset is better, but once the number of training images surpasses $\app$100M, the larger 17k hashtag dataset prevails, presumably because it represents a more diverse visual distribution with more fine-grained concepts.

\paragraph{3.1.3~~What is the effect of hashtag label noise on model accuracy?} A major difference between hashtag supervision and the labels provided in datasets such as ImageNet is that hashtag supervision is inherently \emph{noisy}: users may apply hashtags that are irrelevant to the visual content of the image, or they may have left out hashtags that would have been visually relevant \cite{misra2016bias}. Because an exact characterization of this label noise is difficult, instead, we investigate the effect of injecting additional label noise on the accuracy of our networks. To do so, we pretrain ResNeXt-101 32$\times$16d networks on a version of IG-1B-17k in which we randomly replaced $p\%$ of the hashtags by hashtags obtained by sampling from the marginal distribution over hashtags (excluding the tag to be replaced). 

\begin{figure}[t!]
\begin{floatrow}

\ffigbox{
  \centering
  \includegraphics[width=\linewidth]{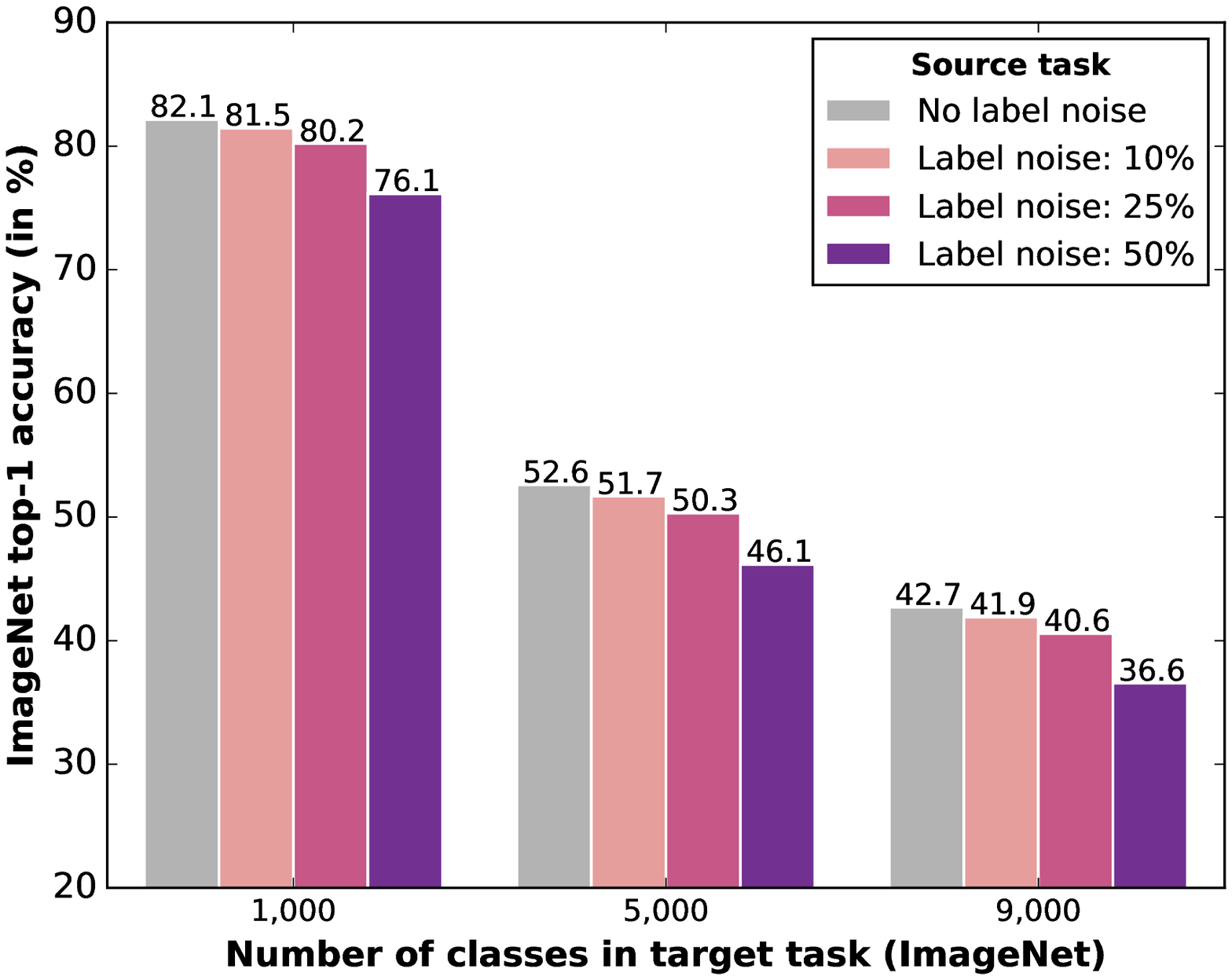}
}{
  \caption{\small Classification accuracy of ResNeXt-101 32$\times$16d, pretrained on IG-1B-17k, on val-IN-\{1k, 5k, 9k\} at three levels of injected label noise. The no-label-noise baseline is trained on the original hashtags. Only the linear classifier is trained on the target task.}\label{fig:figure3}
}

\ffigbox{
  \centering
  \includegraphics[width=\linewidth]{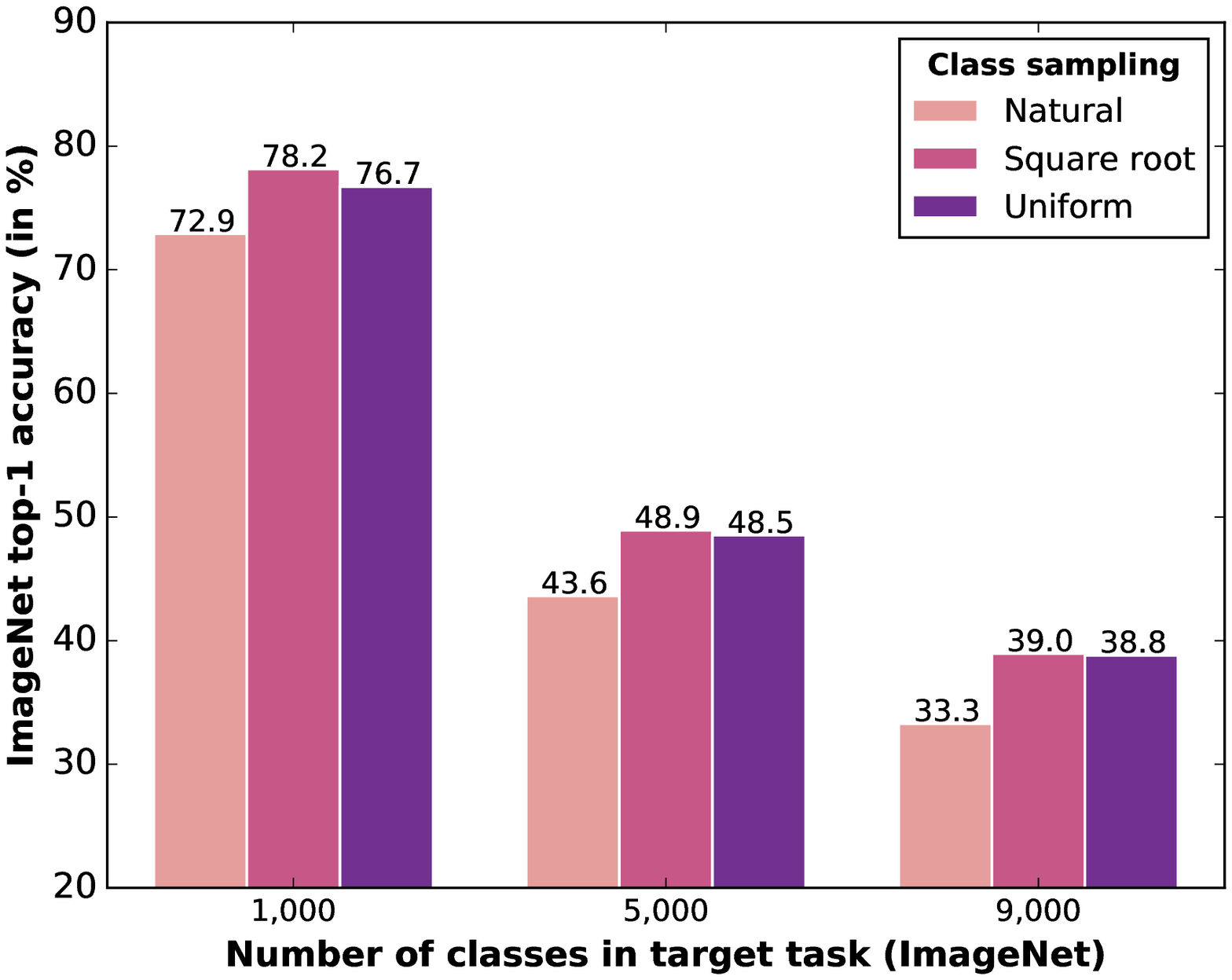}
 }{
   \caption{\small Classification accuracy of ResNeXt-101 32$\times$4d, pretrained on IG-1B-17k, on val-IN-\{1k, 5k, 9k\} for three different hashtag sampling strategies: natural sampling, uniform sampling, and square-root sampling. Only the linear classifier is trained on the target task.}\label{fig:figure4}
}

\end{floatrow}
\end{figure}

Figure~\ref{fig:figure3} shows the ImageNet classification accuracy of the resulting networks for different numbers of classes at three levels, $p$, of artificial label noise as well as for a baseline in which no artificial label noise was added during pretraining. We only train the final linear classifier on the target task, because full finetuning may mask the damage caused by pretraining noise. \emph{The results suggest that the networks are remarkably resilient against label noise:} a noise level of $p\!=\!10\%$ leads to a loss of less than $1\%$ in classification accuracy, and at $p \!=\! 25\%$ label noise, the reduction in accuracy is around $2\%$. These results suggest that label noise may be a limited issue if networks are trained on billions of images.

\paragraph{3.1.4~~How does the sampling of pretraining data impact accuracy?} Another difference between hashtag and ImageNet supervision is that, like in language modeling, hashtags are governed by a Zipfian distribution. Prior studies in language modeling found that resampling Zipfian distributions reduces the impact of the head of the word distribution on the overall training loss \cite{mikolov2013word2vec}. Motivated by this work, we perform experiments in which we evaluate three different types of data sampling in the Instagram pretraining: (1) a \emph{natural} sampling in which we sample images and hashtags according to the distribution by which they appear on Instagram; (2) \emph{square-root} sampling \cite{mikolov2013word2vec} in which we take the square-root of the head of the hashtag distribution, renormalize, and sample according to the resulting distribution (due to practical considerations, our implementation is slightly different; see supplemental material); and (3) \emph{uniform} sampling in which we sample a hashtag uniformly at random, and then sample an image that has this hashtag associated to it uniformly at random \cite{joulin2016learning}. (Aside from this experiment, we always pretrain on Instagram data using square-root sampling.) As before, we only train the final linear classifier on the target task.

Figure~\ref{fig:figure4} displays classification accuracy as a function of the number of ImageNet classes for networks that were pretrained on IG-1B-17k using the three sampling strategies. The results show that resampling of the hashtag distribution is important in order to obtain good transfer to ImageNet image-classification tasks: using uniform or square-root sampling leads to an accuracy improvement of $5$ to $6\%$ irrespective of the number of ImageNet classes in the transfer task. In line with prior results, the figure also shows that larger hashtag vocabularies lead to increasing accuracy improvements as the number of target classes grows.

\paragraph{3.1.5~~With billions of images, is transfer learning model-capacity bound?}
\begin{wrapfigure}[17]{R}{0.5\textwidth}
  \centering
  \vspace{-20pt}
  \includegraphics[width=\linewidth]{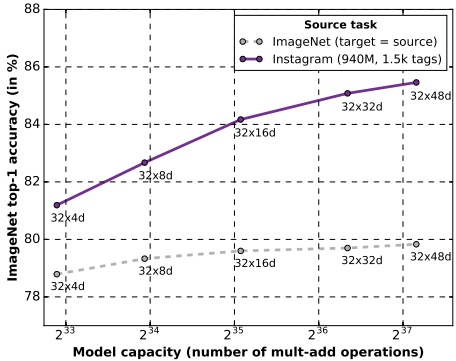}
  \caption{\small Classification accuracy on val-IN-1k using ResNeXt-101 32$\times$\{4, 8 16, 32, 48\}d with and without pretraining on the IG-940M-1.5k dataset.}
  \vspace{-5pt}
  \label{fig:figure9}
\end{wrapfigure}
Now, we look at what happens when we train convolutional networks that are substantially larger than those typically used in recent studies (and our experiments so far). In particular, we use IG-940M-1.5k to pretrain ResNeXt-101 32$\times$32d and ResNeXt-101 32$\times$48d, which have 2.4$\times$ and 4.3$\times$ more add-mult FLOPs than ResNeXt-101 32$\times$16d, respectively. Using these ``super-sized'' models improves val-IN-1k results over the 32$\times$16d model from 84.2\% top-1 accuracy to 85.1\% and 85.4\%, respectively (top-5 accuracy: from 97.2\% to 97.5\% and 97.6\%). By comparison, when training from scratch on IN-1k, top-1 accuracy saturates at around 79.6\% with the 32$\times$16d model and does not meaningfully increase by using larger models. These results, plotted in Figure~\ref{fig:figure9}, indicate that with large-scale Instagram hashtag training, transfer-learning performance appears bottlenecked by model capacity.

\begin{figure}[t!]
\begin{floatrow}
\ffigbox{
  \centering
  {\includegraphics[width=\linewidth]{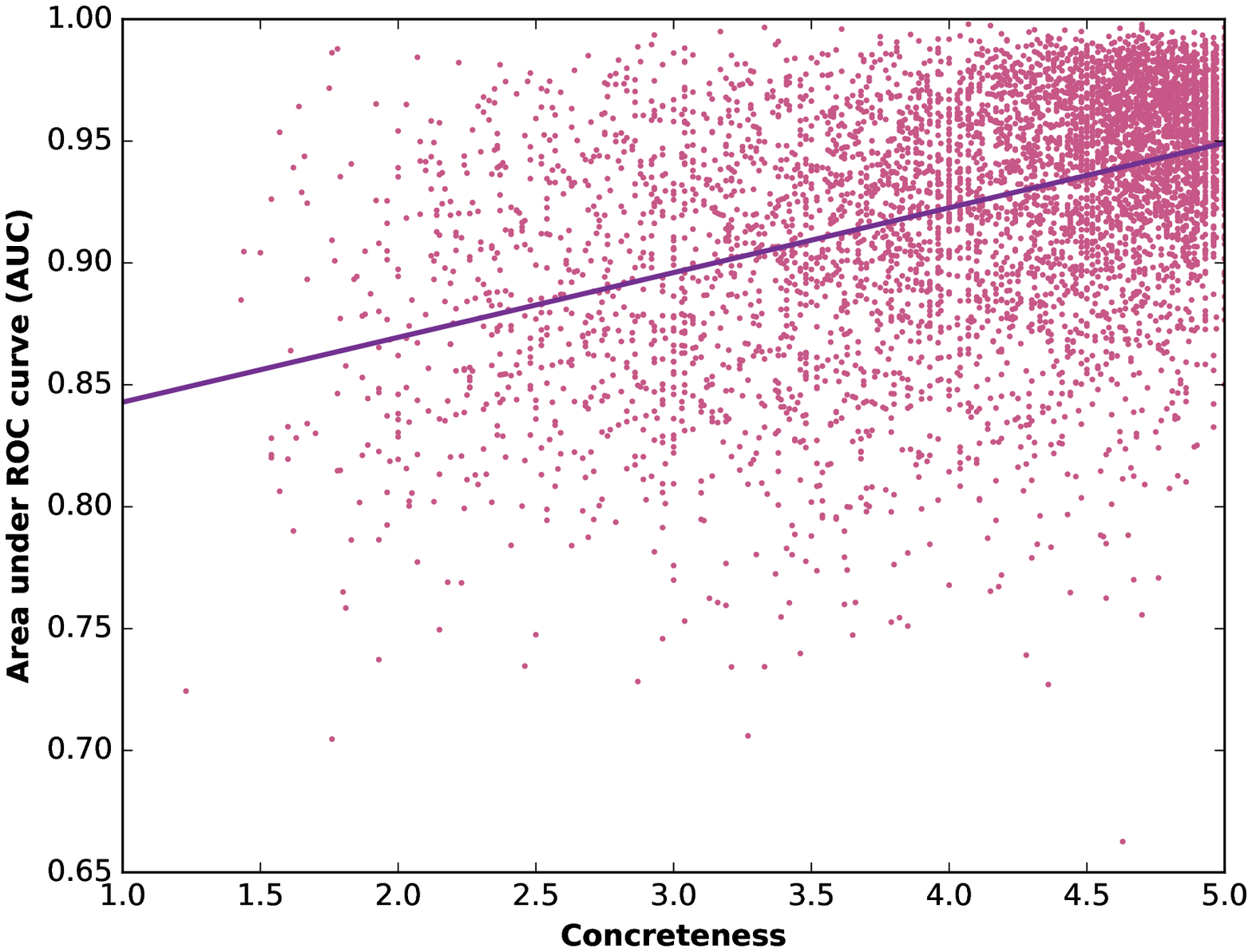}}\vspace{-0.5em}
}{
  \caption{\small Area under ROC curve (AUC) for hashtag prediction as a function of the hashtag concreteness \cite{brysbaert2014concreteness}, and corresponding least-squares fit ($\rho \!=\! 0.43$).}\label{fig:figure6}
}
\ffigbox{
  \centering
  {\includegraphics[width=.95\linewidth]{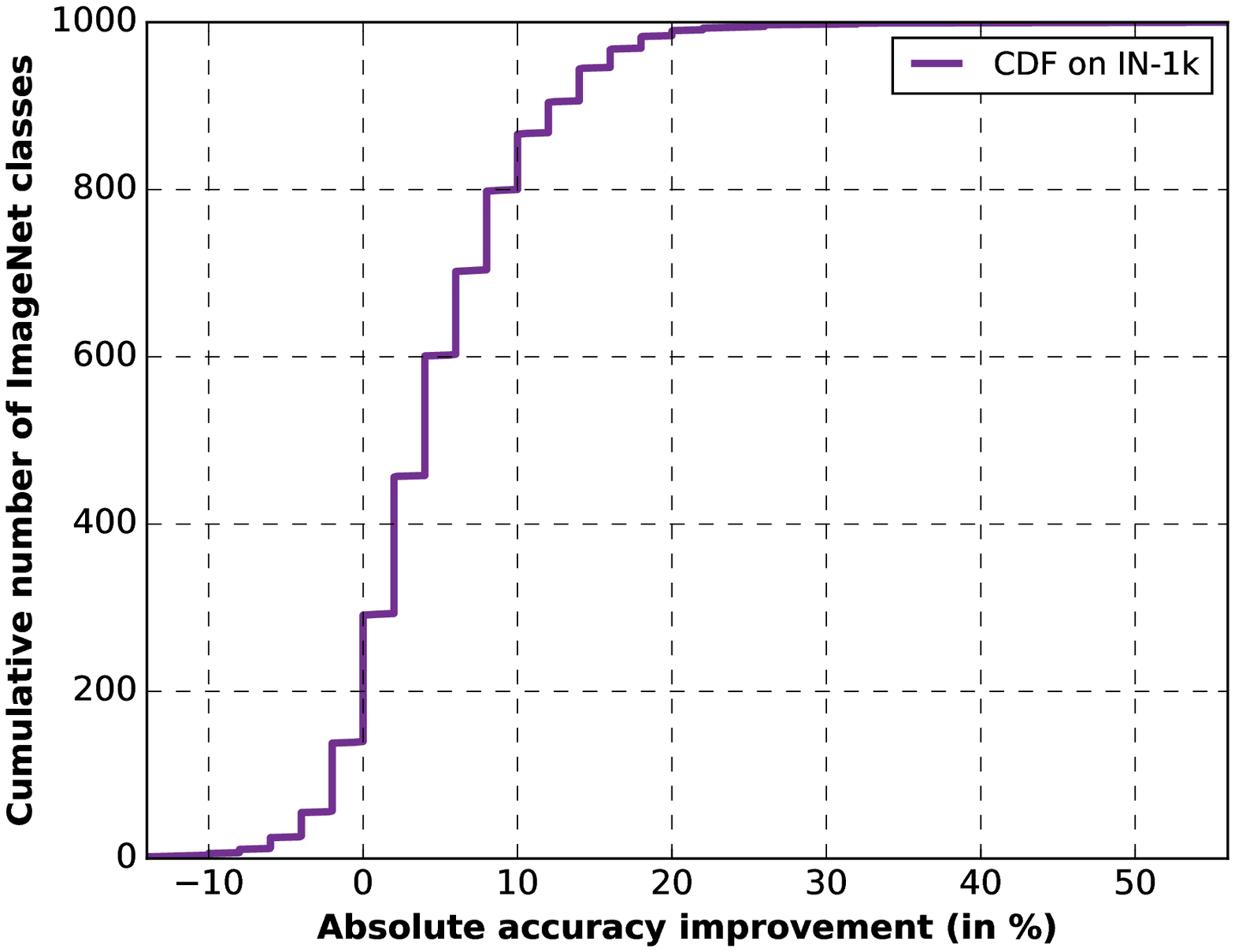}}\vspace{-0.5em}
}{
  \caption{\small CDF of absolute per-class accuracy improvements on IN-1k validation set of an Instagram-pretrained network compared to ImageNet baseline.}\label{fig:figure7}
}
\end{floatrow}
\end{figure}

\paragraph{3.1.6~~On what visual classes is Instagram pretraining most helpful?} Our results with different hashtag vocabularies suggest that choosing the right hashtag vocabulary may be at least as important as scaling model training to billions of images. Specifically, we expect that some hashtags are easier to predict because they are more ``visually concrete'' than others: whereas \texttt{\#eiffeltower} corresponds to a very specific visual scene, \texttt{\#party} may correspond to a large variety of visual scenes. We matched the 17k Instagram hashtags with a list of 40k ``concreteness'' values of nouns \cite{brysbaert2014concreteness} to obtain 5,730 hashtag concreteness values. Figure~\ref{fig:figure6} displays these hashtag concreteness values and the accuracy of predicting the hashtags correctly (measured in terms of AUC on balanced validation sets) in a scatter plot. The figure suggests a clear relation between the concreteness of a noun and the model's ability to predict the corresponding hashtag: the Pearson correlation, $\rho$, between both variables is $0.43$.

We also analyze the effect of Instagram pretraining on recognizing individual IN-1k classes. Figure~\ref{fig:figure7} shows the cumulative distribution function (CDF) of the absolute accuracy improvement per class of a ResNeXt-101 32$\times$16d network pretrained on IG-3.5B-17k compared to the same network trained on IN-1k. Accuracy improves on more than 80\% of the IN-1k classes, with 20\% of classes gaining at least 10 percentage points.

\subsection{Object Detection}

\begin{figure}[t!]
  \centering
  \subfloat[{\plottitle ~~~~~~~Target task: COCO detection (box AP)}]{\includegraphics[width=0.5\linewidth]{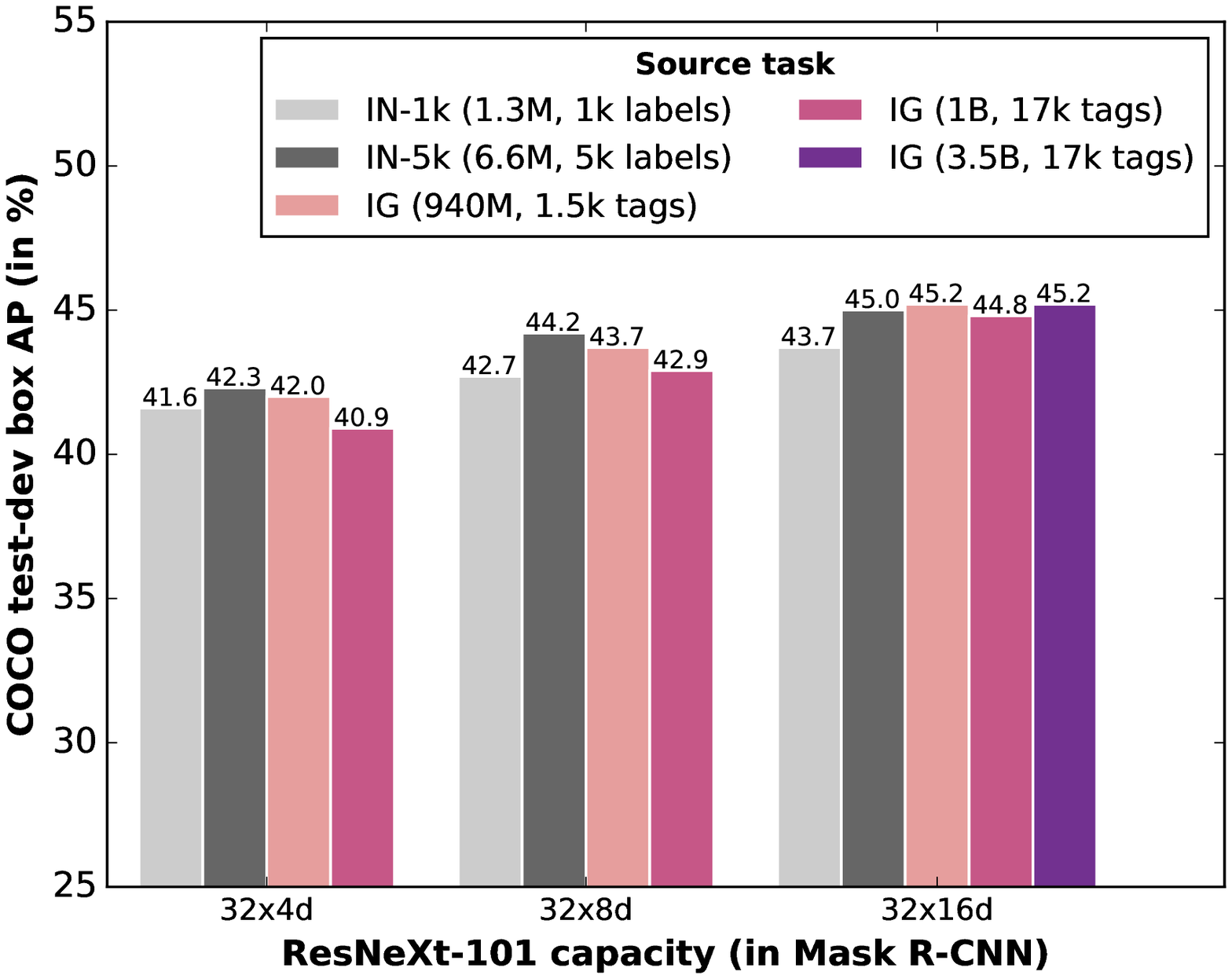}}
  \subfloat[{\plottitle ~~~~~~~Target task: COCO detection (box AP@50)}]{\includegraphics[width=0.5\linewidth]{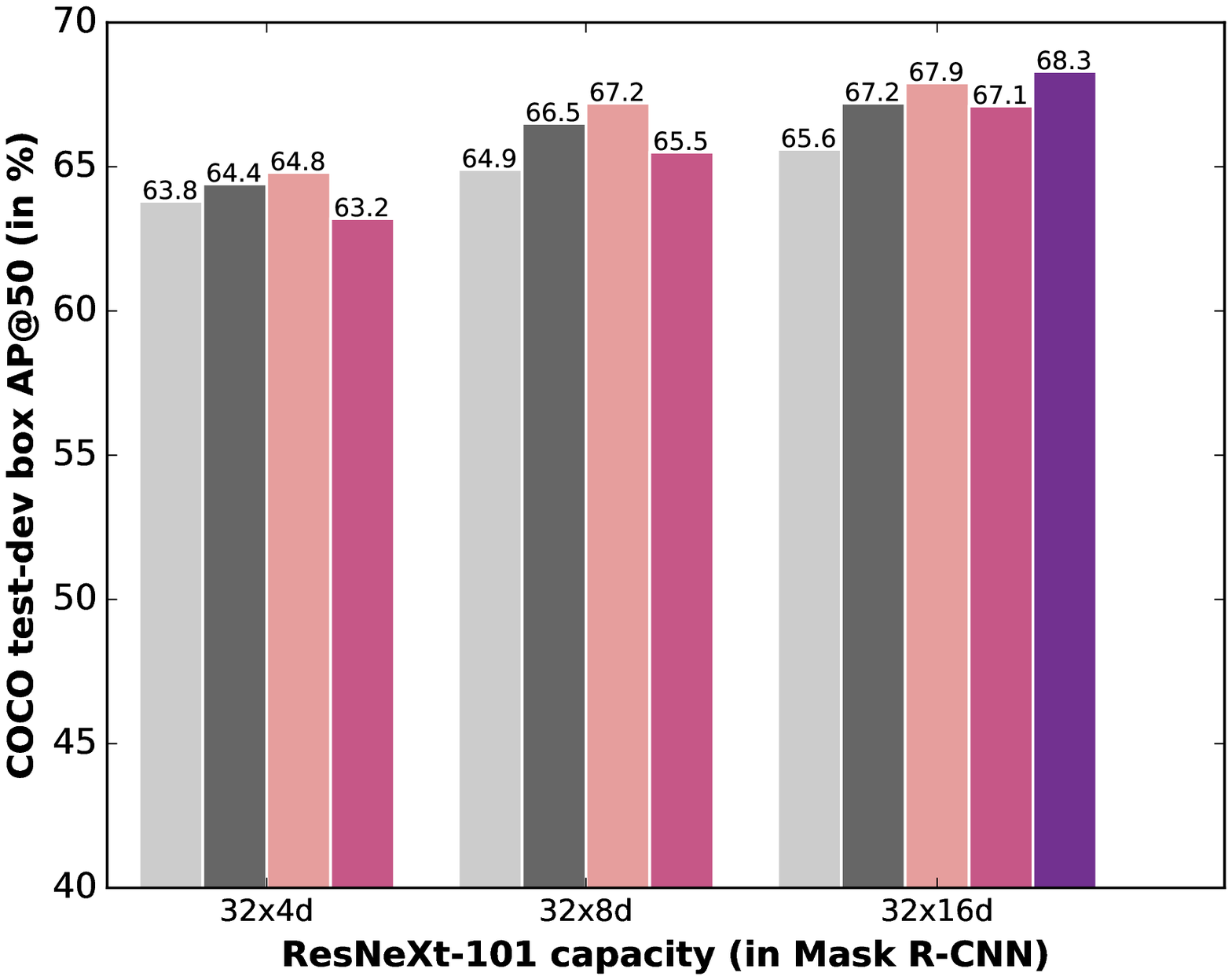}}\\
  \vspace{-0.5em}
  \subfloat[{\plottitle ~~~~~~~Target task: COCO detection (mask AP)}]{\includegraphics[width=0.5\linewidth]{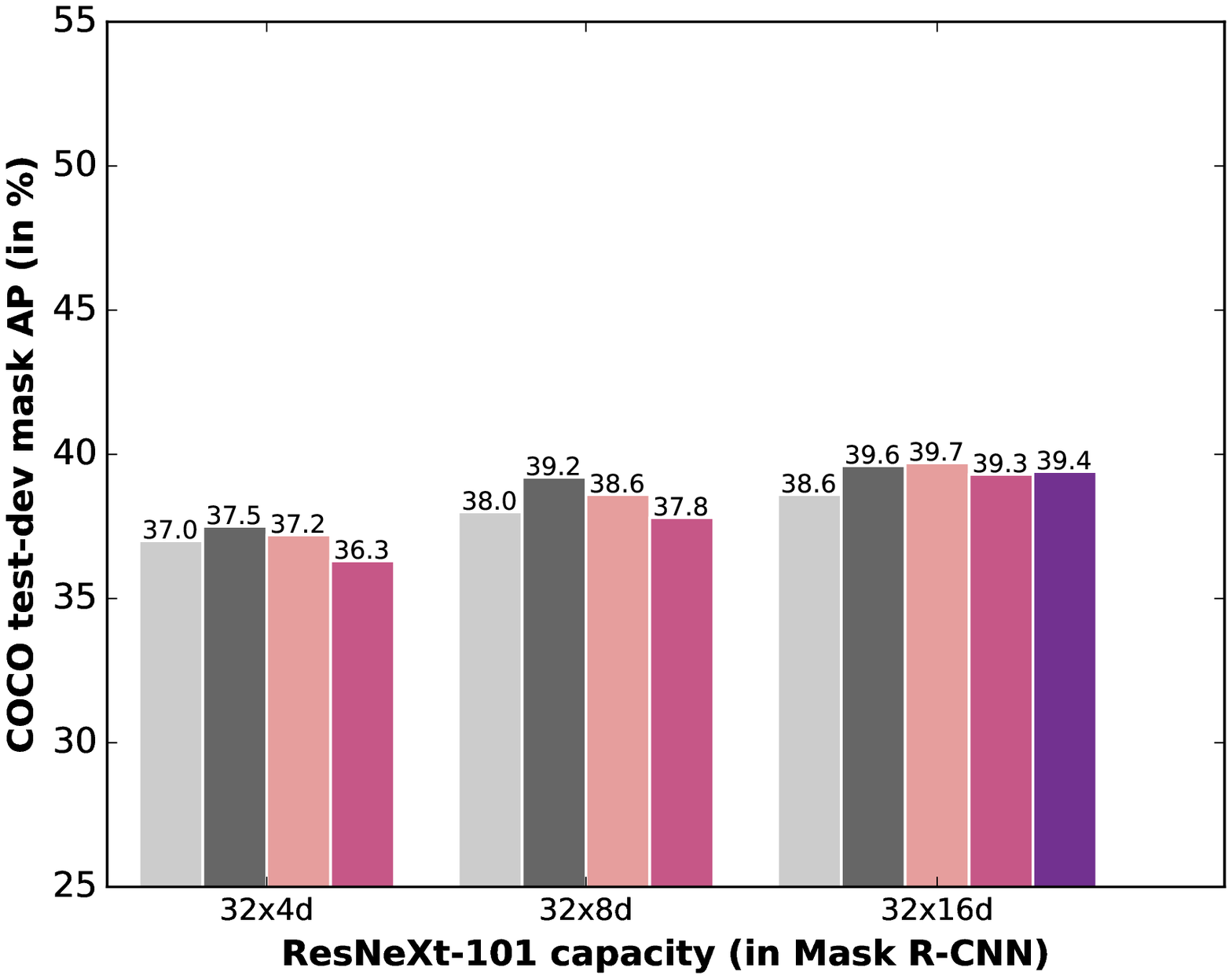}}
  \subfloat[{\plottitle ~~~~~~~Target task: COCO detection (mask AP@50)}]{\includegraphics[width=0.5\linewidth]{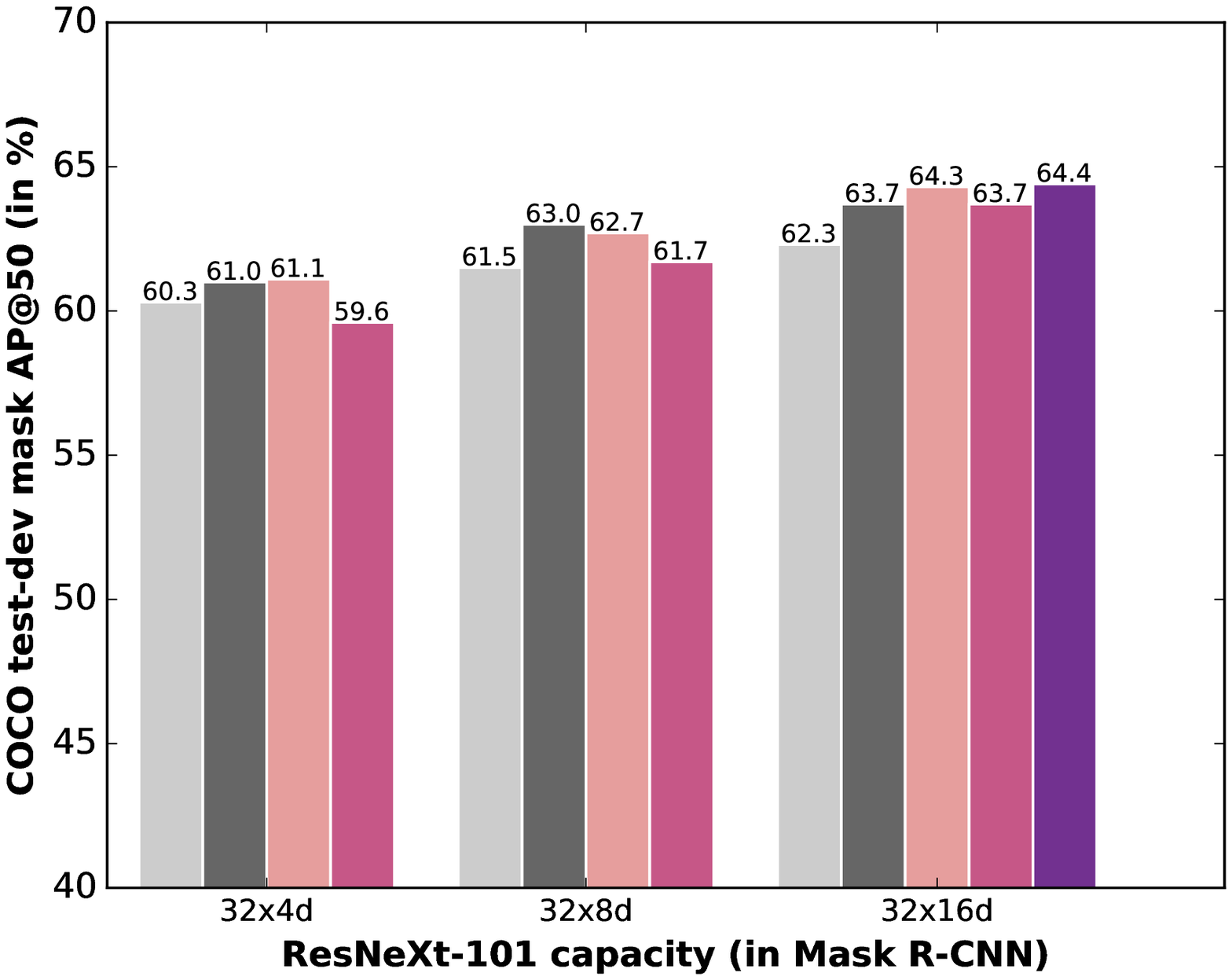}}
  \vspace{-0.5em}
  \caption{\small Transfer to object detection and instance segmentation with Mask R-CNN. We compare ResNeXt-101 FPN backbones of three different capacities using a variety of source pretraining tasks. Higher is better.}
\label{fig:figure8}
\end{figure}

We have looked at target tasks that require image classification, but we are also interested in observing if pretraining on Instagram hashtag data can improve object detection and instance segmentation tasks by finetuning networks on the COCO dataset \cite{mscoco2014}. We use Mask R-CNN \cite{he2017mask,Detectron2018} and experiment with ResNeXt-101 FPN \cite{Lin2017} backbones of three different capacities (see Figure \ref{fig:figure8}).

We compare performance on the 2017 test-dev set using several different pretrained networks. As baselines, we use IN-\{1k, 5k\} pretraining (IN-9k performs no better than IN-5k) and compare them to IG-940M-1k and IG-1B-17k. For the largest model (32$\times$16d) we also include results for IG-3.5B-17k. We use standard settings  \cite{Detectron2018} for end-to-end Mask R-CNN training with one exception: for the Instagram pretrained models we found it necessary to perform grid search for the finetuning learning rate on the validation set. We found that models pretrained on the Instagram data require finetuning learning rates that are \app 4-10$\times$ lower than ImageNet pretrained models (see supplemental material). This finding illustrates that finetuning recipes developed for ImageNet pretrained models do not transfer to new pretraining sets: a larger amount of pretraining data implies the need for lower finetuning learning rates.

Figure \ref{fig:figure8} shows two interesting trends. First, we observe that \emph{when using large amounts of pretraining data, detection is model capacity bound:} with the lowest capacity model (32$\times$4d), the gains from larger datasets are small or even negative, but as model capacity increases the larger pretraining datasets yield consistent improvements. We need even larger models to take advantage of the large-scale pretraining data. The second trend we observe comes from comparing COCO's default AP metric (average precision averaged over intersection-over-union (IoU) overlap thresholds 0.5:0.95) to AP@50 (average precision computed at IoU threshold 0.5 only). The former emphasizes precise localization while the later allows for looser localization. We observe that the improvement over IN-\{1k, 5k\} pretraining from IG-1B-1k is much larger in terms of AP@50. Thus, the gains from Instagram pretraining may be primarily due to improved object classification performance, rather than spatial localization performance. Further evidence comes from experiments with keypoint detection using Mask R-CNN, where we found that compared to IN-1k pretraining, IG-1B-1k pretraining leads to worse results (65.3\% \vs 67.0\% keypoint AP). These two findings suggest that pretraining for Instagram hashtag classification may reduce spatial localization performance while improving classification.

\section{Related Work}
\label{sec:related_work}

Our study is part of a larger body of work on training convolutional networks on large, weakly supervised image datasets. Sun \etal \cite{sun2017unreasonable} train convolutional networks on the JFT-300M dataset of 300 million weakly supervised images. Our Instagram datasets are an order of magnitude larger than JFT-300M, and collecting them required much less manual annotation work (see Section~\ref{sec:ig_datasets}). Due to the larger training set size and the use of better network architectures, we obtain substantially higher accuracies on transfer tasks: \eg, we obtain 85.4\% top-1 accuracy on ImageNet-1k, compared to $79.2\%$ reported in \cite{sun2017unreasonable}.

Other prior studies \cite{joulin2016learning,li2017visualngram} trained convolutional networks to predict words or n-grams in comments on a collection of 100 million Flickr photos and corresponding comments \cite{thomee2016yfcc100m}. Word or n-gram supervision is weaker than hashtag supervision because it is less structured, as reflected in the relatively poor transfer of features to ImageNet reported in \cite{joulin2016learning}. Other work \cite{veit2017separating} also trained networks to predict hashtags on the Flickr dataset but, unlike our study, does not investigate transfer of the resulting networks to other tasks. In addition to Flickr hashtags, \cite{gross2017hard} trained hard mixture of expert models on food-related Instagram hashtags; our focus is on standard recognition networks and general hashtags. Other studies on hashtag prediction \cite{denton2015hashtag} do not train convolutional networks from scratch, but train linear classifiers to predict relevant hashtags from pre-defined image features. Several other works have trained models on web-scale image data for other purposes, such as face recognition \cite{schroff2015facenet,taigman2015webscale} and similarity search \cite{johnson2017billion,stewenius2012size}, but to the best of our knowledge, we are the first to report the results of experiments that involve training convolutional networks from scratch on billions of images.

\section{Discussion}
\label{sec:discussion}
We have attempted to explore the limits of supervised pretraining. In addition to producing state-of-the-art results on the ImageNet-1k benchmark task (85.4\% single-crop, top-1 accuracy; 97.6\% single-crop, top-5 accuracy) and several other vision tasks, our study has led to four important observations:

\begin{enumerate}
\item Our results suggests that, \emph{whilst increasing the size of the pretraining dataset may be worthwhile, it may be at least as important to select a label space for the source task to match that of the target task}. We found that networks trained on a hashtag vocabulary that was designed to match the classes in the ImageNet-1k dataset outperformed those trained on twice as many images without such careful selection of hashtags (Figure~\ref{fig:figure2}, top left). This observation paves the way for the design of ``label-space engineering'' approaches that aim to optimally select (weakly supervised) label sets for a particular target task. Such label-space engineering may be much more fruitful than further increasing the scale of the data on which models are trained.
\item In line with prior work \cite{joulin2016learning,sun2017unreasonable}, \emph{we observe that current network architectures are underfitting when trained on billions of images}. Whilst such underfitting does lead to very high robustness to noise in our hashtag targets (Figure~\ref{fig:figure3}), our results do suggest that accuracy improvements on target tasks may be obtained by further increases of the capacity of our networks (Figure~\ref{fig:figure2}). Capacity may be increased, for instance, by increasing the number of layers and the number of filters per layer of existing architectures or by mixtures-of-experts \cite{gross2017hard} (using model parallelization across GPUs). However, it is not unthinkable that some of the design choices that were made in current network architectures are too tailored to ImageNet-1k classification, and need to be revisited when training on billions of images with hashtag supervision.
\item Our results also underline the importance of increasing the visual variety that we consider in our benchmark tasks. They show that \emph{the differences in the quality of visual features become much more pronounced if these features are evaluated on tasks with a larger visual variety}. For instance, we find that the accuracy difference between models pretrained using two different vocabularies increases as the number of target classes increases (Figure~\ref{fig:figure2}): if we would have only evaluated our models on ImageNet-1k, we would have concluded they learned visual features of similar quality, whereas results on ImageNet-9k show that one model learns substantially better features than the other. We believe evaluation on more ImageNet classes is a good step towards a more comprehensive assessment of visual recognition models.
\item Results from transferring our models to object detection, instance segmentation, and keypoint detection tasks suggestion that \emph{training for large-scale hashtag prediction improves classification while at the same time possibly harming localization performance}. This opens a future direction of modifying large-scale, weakly supervised pretraining tasks to better suit the localization needs of important target tasks like detection and human pose estimation.
\end{enumerate}
In closing, we reflect on the remarkable fact that training for hashtag prediction, \emph{without the need for additional manual annotation or data cleaning}, works at all. We believe our study illustrates the potential of natural or ``wild'' data compared to the traditional approach of manually designing and annotating datasets.

\section*{Acknowledgements}
We would like to thank Matthijs Douze, Aapo Kyrola, Andrew Dye, Jerry Pan, Kevin Wilfong, and Martin Englund for helpful discussions and support.

\bibliographystyle{splncs}
\bibliography{uru}

\appendix
\section{Supplemental Material}
\subsection{Hashtag Selection}
All of our Instagram datasets are subsets of $\ig$, where each image in $\ig$ is from a public Instagram post that has at least one hashtag in its caption (we do not consider hashtags from other sources, such as comments). Let $\ighashtags$ be the set of all unique hashtags associated with the images in $\ig$. To construct a dataset, we use a simple data collection pipeline: (1) We select a set of hashtags that is a subset of $\ighashtags$. (2) We randomly samples images from $\ig$ that are tagged with at least one of these selected hashtags. (3) Then, because multiple hashtags may refer to the same underlying concept, we apply a simple process (described below) that utilizes WordNet \cite{wordnet} synsets to merge some hashtags into a single canonical form (\eg, \texttt{\#brownbear} and \texttt{\#ursusarctos} are merged). (4) Finally, for each sampled image, we replace each of its hashtags with its canonical form and discard any hashtags that were not in the selected set. The canonical hashtags are used as labels for training and evaluation.

\paragraph{Hashtag-to-synset matching.} Rather than taking a random subset of hashtags from $\ighashtags$ in step (1) above, we start with a set of WordNet synsets $\synsets$ and filter the hashtags in $\ighashtags$ by accepting only the ones that \emph{match} to any synset in $\synsets$. To determine if a match exists, we define a function $s(\synsets, h)$ that returns the (possibly empty) subset of $\synsets$ that matches hashtag $h \in \mathcal{H}$. The function $s$ is implemented using the \texttt{nltk} interface to WordNet. Specifically, $s$ returns the union of synsets found by querying WordNet using calls to \texttt{nltk.corpus.wordnet.synsets($x$)} for different values of $x$, which are query strings derived from $h$. As values of $x$, we use the original hashtag $h$ as well as all bigrams formed by inserting a space character at each position in $h$.

\paragraph{Canonical hashtag merging.} Our hashtag merging process in step (3) is also based on WordNet synsets. We consider two hashtags $h \in \mathcal{H}$ and $h' \in \mathcal{H}$ duplicates if and only if $s(\synsets, h) = s(\synsets, h')$. Herein, $s(\synsets, h)$ is the function defined above that returns the set of all synsets that $h$ matches in $\synsets$. For hashtag merging, we set $\synsets$ to all WordNet synsets. This conservative deduplication strategy merges two hashtags whenever they exactly coincide in all possible word senses.

\subsection{Image Deduplication}
\label{Image Deduplication}
We implemented a two-stage deduplication procedure. In the first stage, we search the set of 3.5 billion Instagram images\footnote{This is the largest data set we used and all other image sets are subsets of this one.} using an approximate nearest neighbor search algorithm to identify 128 potential duplicates for each query image (\eg, an image from val-IN-1k). In the second stage, we compute high-quality image features for these duplicate candidates, compute pairwise distances between the candidates and the query image, and apply a conservative distance threshold to generate potential pairs of duplicates for annotation by human annotators. We describe both stages separately below.

In the \textbf{first stage}, we remove the last five\footnote{The hyperparameters of our deduplication pipeline were manually tuned using preliminary experiments on the Holidays \cite{jegou2018holidays} dataset.} convolutional layers from a ResNet-50 model. We resize each of the Instagram images such that its height or width (whichever is larger) is $400$ pixels, and perform a forward-pass through the truncated convolutional network. We compute regional maximum activations of convolutions (R-MAC) features \cite{gordo2016rmac,tolias2016rmac} from the resulting feature maps: \ie, we perform max-pooling on fixed sub-regions of the feature maps. R-MAC features substantially improve the performance of the deduplication method: they add invariance to cropping transformations that may have been applied on the images. We whiten the 2048-dimensional R-MAC features using PCA to obtain a 512-dimensional descriptor for each image. The resulting image descriptor is scalar-quantized to 8 bits per dimension to facilitate storage on disk. 

To construct a searchable index of 3.5 billion images, we undo the scalar quantization, L2-normalize the resulting vector, and apply optimized product quantization (OPQ; \cite{ge2013opq}) to further reduce the feature representation to 256 dimensions. Next, we apply a coarse product quantizer \cite{jegou2011pq} with 2 sub-quantizers that each operate on 128 of the 256 dimensions. Each sub-quantizer uses k-means clustering to quantize the 128 dimensions into 14 bits. The resulting 28-bit image representation is used to construct an inverted list of the 3.5 billion images. Finally, we apply a product quantizer with 32 sub-quantizers on the residual feature vector; each of the sub-quantizers uses k-means to quantize $256 / 32 = 8$ dimensions of the image representation into $8$ bits. The resulting 32-byte representation is stored with the image in the corresponding inverted list. All preprocessors and quantizers were trained on 2 million images; we implemented the image index using Faiss \cite{johnson2017billion}.

Given a query image from the validation set of a target task, we search the resulting index as described in \cite{jegou2011pq}. First, we compute R-MAC features for the query image, and preprocess the features using PCA, scalar quantization, and OPQ. Next, we apply the coarse quantizer on the resulting descriptor and find the $256$ nearest sub-quantizers (in terms of Hamming distance). We compute the residual for each of these sub-quantizers, and compute the squared distances between this residual and the residuals stored in the corresponding entries in the inverted list. This produces distance estimates, which we use to select the $128$ nearest neighbors (in terms of the distance map) efficiently using a max-heap implementation.

In the \textbf{second stage}, we compute R-MAC features for the query image and each of the $128$ identified neighbors in the same way as in the first stage, however, in this stage we do not compress the $2048$-dimensional R-MAC features in any way. We compute exact squared Euclidean distances between the features of the query and neighbor images, and apply a (very conservative) distance threshold of $0.6$ on each of the distances. For each image in the query set (\eg, an image from val-IN-1k) that has at least one neighbor in IG-3.5B for which the R-MAC feature distance is smaller than this threshold, we manually annotated the $21$ nearest neighbors (in terms of the R-MAC feature distance) to assess whether or not the query image has duplicates in the Instagram dataset. 

This procedure led us to $150$ val-IN-50k-1k (0.30\%), $10$ val-CUB-6k-200 (0.17\%), $151$ val-Places-37k-365 ($0.41\%$), and $6$ val-COCO-5k-80 ($0.12\%$) images as duplicates. In the results in the main paper, we included these duplicates in the reported accuracies. In Table~\ref{tab:lower_bound}, we also report a conservative lower bound on the accuracy of our best models that treats all images that have duplicates in the training set as being classified incorrectly\footnote{We note that similar lower bounds should also be computed for networks that are pretrained on ImageNet and then transferred to other datasets (for instance, it is estimated that approximately $5\%$ of the test images in val-CUB-6k-200 are also in train-IN-1M-1k), but we did not perform the required duplicate analyses.}.

\begin{table*}[h!]
  \caption{\small \textbf{Lower bounds on reported accuracies.} Measured top-1 accuracies (left column) and conservative lower bounds on those accuracies obtained by considering all test images with duplicates in the training set as being classified incorrectly (right column). See text in Section~\ref{Image Deduplication} for details.}
\resizebox{0.6\linewidth}{!}{
{\tablestyle{4pt}{1.1}
\begin{tabular}{lp{26mm}p{26mm}}
\toprule
  \bf Dataset & \bf Measured \newline Accuracy (\%) & \bf  Lower-bound \newline Accuracy (\%) \\
\midrule
  val-IN-50k-1k & 84.2 & 83.9 \\
  val-CUB-6k-200 & 89.2 & 89.0 \\
  val-Places-37k-365 & 58.0 & 57.6 \\
\bottomrule
\end{tabular}}
}
\label{tab:lower_bound}
\end{table*}

\subsection{Training Details}
\paragraph{Classification hyperparameters.} Our training hyperparameters match those in \cite{goyal2017accurate}. We briefly summarize them here for completeness. For SGD, we use Nesterov momentum \cite{Nesterov2004} of 0.9 and weight decay of 0.0001. Weight decay is not applied to the batch normalization (BN) scale ($\gamma$) and bias ($\beta$) parameters. All convolutional filters are initialized according to \cite{He2015}. Weights of the final fully connected layer are sampled from a zero-mean Gaussian with standard deviation 0.01. Following \cite{goyal2017accurate}, the final BN scale parameter $\gamma$ of each residual block is initialized to 0 (instead of 1). We use standard image rescaling and data augmentation as described in \cite{goyal2017accurate}. These settings are consistent across ImageNet and Instagram pretraining.

Our feature transfer experiments require training an L2-regularized logistic regressor. Since the objective function is convex, optimization is straightforward and we only need to perform a grid search for the L2 penalty. For train-CUB-200, the optimal penalty was 0.001; for all other datasets it was 0.0001. When training from scratch or performing full network finetuning, the training hyperparameters are more complex. We give details in Tables \ref{tab:sched:scratch} and \ref{tab:sched:finetuning}.

\begin{table*}[h!]
  \caption{\small \textbf{Training schedules for pretraining of models.} For ImageNet datasets, the total training length is given in terms of dataset epochs. The learning rate (LR) is set to the initial LR at the start of training (the same for all cases, given as a minibatch size normalized reference value \cite{goyal2017accurate}) and decayed by the specified LR decay factor according to the steps, also given in epochs for ImageNet datasets. For Instagram datasets, the total training length is given in terms of the number of images processed. The LR is decayed from the initial value by the specified factor at equally spaced steps; the total number LR decay steps is given in the LR steps column. We specify the schedules for the two dataset extremes for the cases of 1.5k and 17k hashtags. When training on a number of images between the two extremes, we linearly interpolate the training schedule.}
\resizebox{0.98\linewidth}{!}{
{\tablestyle{4pt}{1.1}
\begin{tabular}{llllll}
\toprule
  \bf Dataset & \bf Total length & \bf LR steps & \bf Initial LR & \bf LR decay & \bf Weight decay \\
\midrule
  train-IN-1k & 100 epochs & [30, 30, 30, 10]         & 0.1/256 & 0.1 & $\scinum{1}{-4}$ \\
  train-IN-5k & 40 epochs  & [15, 15, 6, 2]           & 0.1/256 & 0.1 & $\scinum{1}{-4}$ \\
  train-IN-9k & 25 epochs  & [9.35, 9.35, 3.75, 1.25] & 0.1/256 & 0.1 & $\scinum{1}{-4}$ \\
\hline
  train-IG-940M-1.5k & $\scinum{1925}{6}$ images & 20 & 0.1/256 & 0.5          & $\scinum{1}{-4}$ \\
  train-IG-940k-1.5k & $\scinum{300}{6}$ images  & 20 & 0.1/256 & 0.5          & $\scinum{1}{-4}$ \\
  train-IG-3.5B-17k  & $\scinum{7000}{6}$ images & 40 & 0.1/256 & $\sqrt{0.5}$ & $\scinum{1}{-4}$ \\
  train-IG-3.5M-17k  & $\scinum{300}{6}$ images  & 20 & 0.1/256 & 0.5          & $\scinum{1}{-4}$ \\
\bottomrule
\end{tabular}}
}
\label{tab:sched:scratch}
\end{table*}

\begin{table*}[h!]
  \caption{\small \textbf{Training schedules for finetuning of models.} For transfer learning with full network finetuning, we used a proper validation set held out from the training set of the target task. Using this validation set, we did a coarse grid search to find the initial LR (chosen from 0.0025/256, 0.00025/256, or 0.000025/256) and the weight decay (chosen from 0.01, 0.001, or 0.0001). In all cases, we fixed the length of the finetuning schedule based on some preliminary experiments.}
\resizebox{0.99\linewidth}{!}{
{\tablestyle{4pt}{1.1}
\begin{tabular}{lllllll}
\toprule
  \bf Source dataset & \bf Target dataset & \bf Total length & \bf LR steps & \bf Initial LR & \bf LR decay & \bf Weight decay \\
\midrule
  train-IG-940M-1.5k & train-IN-1k & 30 epochs & [10, 10, 10] & 0.00025/256 & 0.1 & $\scinum{1}{-4}$ \\
  train-IG-1B-17k    & train-IN-1k & 30 epochs & [10, 10, 10] & 0.00025/256 & 0.1 & $\scinum{1}{-4}$ \\
  train-IG-3.5B-17k  & train-IN-1k & 30 epochs & [10, 10, 10] & 0.00025/256 & 0.1 & $\scinum{1}{-4}$ \\
\hline
  train-IG-940M-1.5k & train-IN-5k & 30 epochs & [10, 10, 10] & 0.0025/256 & 0.1 & $\scinum{1}{-4}$ \\
  train-IG-1B-17k    & train-IN-5k & 30 epochs & [10, 10, 10] & 0.00025/256 & 0.1 & $\scinum{1}{-4}$ \\
  train-IG-3.5B-17k  & train-IN-5k & 30 epochs & [10, 10, 10] & 0.00025/256 & 0.1 & $\scinum{1}{-4}$ \\
\hline
  train-IG-940M-1.5k & train-IN-9k & 24 epochs & [8, 8, 8] & 0.0025/256 & 0.1 & $\scinum{1}{-4}$ \\
  train-IG-1B-17k    & train-IN-9k & 24 epochs & [8, 8, 8] & 0.00025/256 & 0.1 & $\scinum{1}{-4}$ \\
  train-IG-3.5B-17k  & train-IN-9k & 24 epochs & [8, 8, 8] & 0.00025/256 & 0.1 & $\scinum{1}{-4}$ \\
\hline
  train-IG-940M-1.5k & train-CUB-200 & 300 epochs & [100, 100, 100] & 0.00025/256 & 0.1 & $\scinum{1}{-3}$ \\
  train-IG-1B-17k    & train-CUB-200 & 300 epochs & [100, 100, 100] & 0.00025/256 & 0.1 & $\scinum{1}{-3}$ \\
  train-IG-3.5B-7k   & train-CUB-200 & 300 epochs & [100, 100, 100] & 0.00025/256 & 0.1 & $\scinum{1}{-3}$ \\
  train-IN-1.3M-1k   & train-CUB-200 & 300 epochs & [100, 100, 100] & 0.0025/256 & 0.1 & $\scinum{1}{-2}$ \\
\hline
  train-IG-940M-1.5k & train-Places-365 & 8 epochs  & [4, 2, 2]    & 0.00025/256 & 0.1 & $\scinum{1}{-4}$ \\
  train-IG-1B-17k    & train-Places-365 & 8 epochs  & [4, 2, 2]    & 0.00025/256 & 0.1 & $\scinum{1}{-4}$ \\
  train-IG-3.5B-7k   & train-Places-365 & 8 epochs  & [4, 2, 2]    & 0.00025/256 & 0.1 & $\scinum{1}{-4}$ \\
  train-IN-1.3M-1k   & train-Places-365 & 18 epochs & [8, 6, 2, 2] & 0.0025/256 & 0.1 & $\scinum{1}{-4}$ \\
\bottomrule
\end{tabular}}
}
\label{tab:sched:finetuning}
\end{table*}

\paragraph{Detection hyperparameters.} For our object detection experiments, we found it necessary to perform grid search for the initial learning rate; using defaults that work well when finetuning from ImageNet-pretrained models worked poorly when finetuning with Instagram-pretrained models. The grid search was performed on the COCO val2017 set, which results in the paper are reported on the test-dev2017 set, which requires submitting results to the COCO evaluation server. The optimal initial learning rates are shown in Table \ref{tab:sched:det_finetuning}.

\begin{table*}[h!]
  \caption{\small \textbf{Training schedules for finetuning of Mask R-CNN for detection on COCO.} All models are trained with a minibatch size 8 (images) for a total of 180k iterations. The learning rate is decreased by a factor of 0.1 at 120k and 160k iterations.}
\resizebox{0.50\linewidth}{!}{
{\tablestyle{4pt}{1.1}
\begin{tabular}{lll}
\toprule
  \bf Backbone model & \bf Source dataset & \bf Initial LR \\
\midrule
  ResNeXt-101 32$\times$4d & IN-1k & 0.01 \\
  ResNeXt-101 32$\times$4d & IN-5k & 0.01 \\
  ResNeXt-101 32$\times$4d & IG-940M-1.5k & 0.0025 \\
  ResNeXt-101 32$\times$4d & IG-1B-17k & 0.0025 \\
\hline
  ResNeXt-101 32$\times$8d & IN-1k & 0.01 \\
  ResNeXt-101 32$\times$8d & IN-5k & 0.01 \\
  ResNeXt-101 32$\times$8d & IG-940M-1.5k & 0.0025 \\
  ResNeXt-101 32$\times$8d & IG-1B-17k & 0.0025 \\
\hline
  ResNeXt-101 32$\times$16d & IN-1k & 0.01 \\
  ResNeXt-101 32$\times$16d & IN-5k & 0.01 \\
  ResNeXt-101 32$\times$16d & IG-940M-1.5k & 0.0025 \\
  ResNeXt-101 32$\times$16d & IG-1B-17k    & 0.0025 \\
  ResNeXt-101 32$\times$16d & IG-3.5B-17k  & 0.00075 \\
\bottomrule
\end{tabular}}
}
\label{tab:sched:det_finetuning}
\end{table*}

\subsection{Data Resampling}
Hashtag frequencies follow a Zipfian distribution. For example, in the 17k hashtag vocabulary, the most frequent hashtag (\texttt{\#fineart}) appears more than 1 million times as often as the least frequent hashtag (\texttt{\#shirtfront}). Training on the natural distribution may not be optimal and therefore we consider two alternative ways to sample: \emph{uniform} and \emph{square root}. In both cases, we compute a replication factor $r(h)$ for each hashtag $h$: $r(h) = \max(1, \phi(t / f(h)))$, where $\phi(x) = x$ in the case of uniform sampling and $\phi(x) = \sqrt{x}$ in the case of square root sampling. Given an image $I$ with (possibly multiple) hashtags $\{h_i\}$, the image-level replication factor for $I$ is computed as $r(I) = \max_i r(h_i)$. For a set of $n$ unique images, a list of training images is constructed by computing the replication factor for each image, duplicating the image the prescribed number of times, and then randomly permuting the list.\footnote{When an image with multiple hashtags is replicated $r(I)$ times, each individual hashtag $h_i$ is removed as needed such that $h_i$ is only replicated $r(h_i)$ times.} The threshold $t$ is selected such that the final list has a target length matching the desired training schedule length (\eg, processing 2 billion images during training).

\subsection{Comparison with the State of the Art on ImageNet-1k}

\begin{table*}[h!]
  \caption{\small \textbf{Comparison with the state of the art on the ImageNet-1k validation set.} Result table adopted from Zoph \etal \cite{zoph2017transferable}, to which we append the result of ResNeXt-101 32$\times C$d, with $C \in \{16, 32, 48\}$ pretrained on Instagram hashtag data and finetuned on train-IN-1k. All results are based on a single image crop of the specified size (squared). Our results demonstrate that pretraining on billions of images using their hashtags as labels significantly improve on the state-of-the-art ImageNet-1k results, particularly in the case of top-5 accuracy.}
\resizebox{0.99\linewidth}{!}{
{\tablestyle{4pt}{1.1}
\begin{tabular}{llllll}
\toprule
  \bf Model & \bf Image size & \bf Parameters & \bf Mult-adds & \bf Top-1 Acc. (\%) & \bf Top-5 Acc. (\%) \\
\midrule
  Inception V2 \cite{Ioffe2015} & 224 & 11.2M & 1.94B & 74.8 & 92.2 \\
  NASNet-A (5 @ 1538) \cite{zoph2017transferable} & 299 & 10.9M & 2.35B & 78.6 & 94.2 \\
\hline
  Inception V3 \cite{szegedy2016inceptionv3} & 299 & 23.8M & 5.72B & 78.0 & 93.9 \\ 
  Xception \cite{chollet2016xception} & 299 & 22.8M & 8.38B & 79.0 & 94.5 \\
  Inception ResNet V2 \cite{szegedy2016inceptionresidual} & 299 & 55.8M & 13.2B & 80.4 & 95.3 \\
  NASNet-A (7 @ 1920) \cite{zoph2017transferable} & 299 & 22.6M & 4.93B & 80.8 & 95.3 \\
\hline
  ResNeXt-101 64$\times$4 \cite{xie2017resnext} & 320 & 83.6M & 31.5B & 80.9 & 95.6 \\
  PolyNet \cite{zhang2016polynet} & 331 & 92M & 34.7B & 81.3 & 95.8 \\
  DPN-131 \cite{chen2017dual} & 320 & 79.5M & 32.0B & 81.5 & 95.8 \\
  SENet \cite{hu2017squeeze} & 320 & 145.8M & 42.3B & 82.7 & 96.2 \\
  NASNet-A (6 @ 4032) \cite{zoph2017transferable} & 331 & 88.9M & 23.8B & 82.7 & 96.2 \\
\hline
  \emph{Our models:} & & & & & \\
  ~~IG-3.5B-17k ResNeXt-101 32$\times$16d & 224 & 194M & 36B & 84.2 & 97.2 \\
  ~~IG-940M-1.5k ResNeXt-101 32$\times$32d & 224 & 466M & 87B & 85.1 & 97.5 \\
  \bf ~~IG-940M-1.5k ResNeXt-101 32$\times$48d & \bf 224 & \bf 829M & \bf 153B & \bf \bf 85.4 & \bf 97.6 \\
\bottomrule
\end{tabular}}
}
\label{tab:sota}
\end{table*}

\end{document}